%% file: main.tex
\documentclass[10pt,twocolumn,letterpaper]{article}

\usepackage{cvpr}

\makeatletter						%
\@namedef{ver@everyshi.sty}{}
\makeatother
\usepackage{tikz}

\usepackage{times}
\usepackage{epsfig}
\usepackage{graphicx}
\usepackage{amsmath}
\usepackage{amssymb}
\usepackage{booktabs}
\usepackage{algorithm}
\usepackage{algorithmicx}
\usepackage{algpseudocode}
\usepackage{multirow}
\usepackage{array}
\usepackage{subfig}
\usepackage{nopageno}

\usepackage{empheq}
\usepackage[most]{tcolorbox}

\tcbset{highlight math style={colframe=black!60!black,colback=white!50!white,arc=4pt,boxrule=1pt,drop fuzzy shadow}}

\DeclareMathOperator{\E}{E}	
\DeclareMathOperator{\SP}{softplus}	

\usepackage[pagebackref=true,breaklinks=true,letterpaper=true,colorlinks,bookmarks=false]{hyperref}

\cvprfinalcopy 

\def\cvprPaperID{8873} 

\ifcvprfinal\fi
\begin{document}

\title{Adversarial Latent Autoencoders}

\author{Stanislav Pidhorskyi $\quad$ Donald A. Adjeroh  $\quad$ Gianfranco Doretto\\
Lane Department of Computer Science and Electrical Engineering\\
West Virginia University, Morgantown, WV 26506\\
{\tt\small \{ stpidhorskyi, daadjeroh, gidoretto \}@mix.wvu.edu}
\vspace{-4mm}
}

\maketitle

\begin{abstract}

Autoencoder networks are unsupervised approaches aiming at combining generative and representational properties by learning simultaneously an encoder-generator map. Although studied extensively, the issues of whether they have the same generative power of GANs, or learn disentangled representations, have not been fully addressed. We introduce an autoencoder that tackles  these issues jointly, which we call \emph{Adversarial Latent Autoencoder (ALAE)}. It is a general architecture that can leverage recent improvements on GAN training procedures. We designed two autoencoders: one based on a MLP encoder, and another based on a StyleGAN generator, which we call StyleALAE. We verify the disentanglement properties of both architectures. We show that StyleALAE can not only generate $1024\times 1024$ face images with comparable quality of StyleGAN, but at the same resolution can also produce face reconstructions and manipulations based on real images. This makes ALAE the first autoencoder able to compare with, and go beyond the capabilities of a generator-only type of architecture.
\end{abstract}


\input{main_introduction}

\input{main_review}
\input{main_approach}
\input{main_stylealae}
\input{main_experiments1}

\input{main_experiments2}

\vspace{-3mm}
\section{Conclusions}
\vspace{-2mm}

We introduced ALAE, a novel autoencoder architecture that is simple, flexible and general, as we have shown to be efective with two very different backbone generator-encoder networks. Differently from previous work it allows learning the probability distribution of the latent space, when the data distribution is learned in adversarial settings. Our experiments confirm that this enables learning representations that are likely less entangled. This allows us to extend StyleGAN to StyleALAE, the first autoencoder capable of generating and manipulating images in ways not possible with SyleGAN alone, while maintaining the same level of visual detail.

\vspace{-1mm}
\subsubsection*{Acknowledgments}
\vspace{-2mm}

This material is based upon work supported by the National Science Foundation under Grants No. OIA-1920920, and OAC-1761792.

\clearpage

{\small
\bibliographystyle{ieee_fullname}
\bibliography{references}
}

\end{document}

%% file: main_introduction.tex
\section{Introduction}

Generative Adversarial Networks (GAN)~\cite{Goodfellow2014} have emerged as one of the dominant unsupervised approaches for computer vision and beyond. Their strength relates to their remarkable ability to represent complex probability distributions, like the face manifold~\cite{liu2015deep}, or the bedroom images manifold~\cite{Yu2015}, which they do by learning a \emph{generator} map from a known distribution onto the data space. Just as important are the approaches that aim at learning an \emph{encoder} map from the data to a latent space. They allow learning suitable representations of the data for the task at hand, either in a supervised~\cite{Krizhevsky2012,Simonyan14c,Ren2015,he2016deep,Woo2018}, or unsupervised~\cite{Mirza2014,Zhu2017,Higgins2017,Kim2018,Chen2018,Brock2019} manner.

Autoencoder (AE)~\cite{Kingma2014,Rezende2014} networks are unsupervised approaches aiming at combining the ``generative'' as well as the ``representational'' properties by learning simultaneously an \emph{encoder-generator} map. General issues subject of investigation in AE structures are whether they can: (a) have the same generative power as GANs; and, (b) learn disentangled representations~\cite{achille2018emergence}. Several works have addressed (a)~\cite{Makhzani2015,Larsen2016,Donahue2016, Dumoulin2016,Huang2018}. An important testbed for success has been the ability for an AE to generate face images as rich and sharp as those produced by a GAN~\cite{karras2017progressive}. Progress has been made but victory has not been declared. A sizable amount of work has addressed also (b)~\cite{Higgins2017,Kim2018,Eastwood2018}, but not jointly with (a).

We introduce an AE architecture that is general, and has generative power comparable to GANs while learning a less entangled representation. We observed that every AE approach makes the same assumption: \textsl{the latent space should have a probability distribution that is fixed \emph{a priori} and the autoencoder should match it}. On the other hand, it has been shown in~\cite{Karras2019}, the state-of-the-art for synthetic image generation with GANs, that an intermediate latent space, far enough from the imposed input space, tends to have improved disentanglement properties.

The observation above has inspired the proposed approach. We designed an AE architecture where we allow the latent distribution to be learned from data to address entanglement (A). The output data distribution is learned with an adversarial strategy (B). Thus, we retain the generative properties of GANs, as well as the ability to build on the recent advances in this area. For instance, we can seamlessly include independent sources of stochasticity, which have proven essential for generating image details, or can leverage recent improvements on GAN loss functions, regularization, and hyperparameters tuning~\cite{Arjovsky2017, Kurach2018,Miyato2018,Lucic2018,mescheder2018training, Brock2019}. Finally, to implement (A) and (B) we impose the AE reciprocity in the latent space (C). Therefore, we can avoid using reconstruction losses based on simple $\ell_2$ norm that operate in data space, where they are often suboptimal, like for the image space. We regard the unique combination of (A), (B), and (C) as the major techical novelty and strength of the approach. Since it works on the latent space, rather than autoencoding the data space, we named it \emph{Adversarial Latent Autoencoder (ALAE)}.

We designed two ALAEs, one with a multilayer perceptron (MLP) as encoder with a symmetric generator, and another with the generator derived from a StyleGAN~\cite{Karras2019}, which we call \emph{StyleALAE}. For this one, we designed a companion encoder and a progressively growing architecture. We verified qualitatively and quantitatively that both architectures learn a latent space that is more disentangled than the imposed one. In addition, we show qualitative and quantitative results about face and bedroom image generation that are comparable with StyleGAN at the highest resolution of $1024 \times 1024$. Since StyleALAE learns also an encoder network, we are able to show at the highest resolution, face reconstructions as well as several image manipulations based on real images rather then generated.

%% file: main_review.tex
\section{Related Work}

Our approach builds directly on the vanilla GAN architecture~\cite{goodfellow2016nips}. Since then, a lot of progress has been made in the area of synthetic image generation. LAPGAN~\cite{Denton2015} and StackGAN~\cite{Zhang2017,Zhang2018} train a stack of GANs organized in a multi-resolution pyramid to generate high-resolution images. HDGAN~\cite{Zhang2018a} improves by incorporating hierarchically-nested adversarial objectives inside the network hierarchy. In~\cite{Wang2018} they use a multi-scale generator and discriminator architecture to synthesize high-resolution images with a GAN conditioned on semantic label maps, while in BigGAN~\cite{Brock2019} they improve the synthesis by applying better regularization techniques. In PGGAN~\cite{karras2017progressive} it is shown how high-resolution images can be synthesized by progressively growing the generator and the discriminator of a GAN. The same principle was used in StyleGAN~\cite{Karras2019}, the current state-of-the-art for face image generation, which we adapt it here for our StyleALAE architecture. Other recent work on GANs has focussed on improving the stability and robustness of the training~\cite{roth2017stabilizing}. New loss functions have been introduced~\cite{Arjovsky2017}, along with gradient regularization methods~\cite{Nagarajan2017,mescheder2018training}, weight normalization techniques~\cite{Miyato2018}, and learning rate equalization~\cite{karras2017progressive}. Our framework is amenable to these improvements, as we explain in later sections.

Variational AE architectures~\cite{Kingma2014,Rezende2014} have not only been appreciated for their theoretical foundation, but also for their stability during training, and the ability to provide insightful representations. Indeed, they stimulated research in the area of disentanglement~\cite{achille2018emergence}, allowing learning representations with controlled degree of disentanglement between factors of variation in~\cite{Higgins2017}, and subsequent improvements in~\cite{Kim2018}, leading to more elaborate metrics for disentanglement quantification~\cite{Eastwood2018,Chen2018,Karras2019}, which we also use to analyze the properties of our approach. VAEs have also been extended to learn a latent prior different than a normal distribution, thus achieving significantly better models~\cite{tomczak2017vae}.  

A lot of progress has been made towards combining the benefits of GANs and VAEs. AAE~\cite{Makhzani2015} has been the precursor of those approaches, followed by VAE/GAN~\cite{Larsen2016} with a more direct approach. BiGAN~\cite{Donahue2016} and ALI~\cite{Dumoulin2016}  provide an elegant framework fully adversarial, whereas VEEGAN~\cite{Srivastava2017} and AGE~\cite{Ulyanov2018} pioneered the use of the latent space for autoencoding and advocated the reduction of the architecture complexity. PIONEER~\cite{Heljakka2018} and IntroVAE~\cite{Huang2018} followed this line, with the latter providing the best generation results in this category. Section~\ref{sec-relation-to-ae} describes how the proposed approach compares with those listed here.

Finally, we quickly mention other approaches that have shown promising results with representing image data distributions. Those include autoregressive~\cite{VanOord2016} and flow-based methods~\cite{Kingma2018}. The former forego the use of a latent representation, but the latter does not.


%% file: main_approach.tex
\section{Preliminaries}

A Generative Adversarial Network (GAN)~\cite{Goodfellow2014} is composed of a generator network $\mathtt{G}$ mapping from a space $\mathcal{Z}$ onto a \emph{data space} $\mathcal{X}$, and a discriminator network $\mathtt{D}$ mapping from  $\mathcal{X}$ onto $\mathbb{R}$. The $\mathcal{Z}$ space is characterized by a known distribution $p(z)$. By sampling from $p(z)$, the generator $\mathtt{G}$ produces data representing a \emph{synthetic} distribution $q(x)$. Given training data $\mathcal{D}$ drawn from a \emph{real} distribution $p_{\mathcal{D}}(x)$, a GAN network aims at learning $\mathtt{G}$ so that $q(x)$ is as close to $p_{\mathcal{D}}(x)$ as possible. This is achieved by setting up a zero-sum two-players game with the discriminator $\mathtt{D}$. The role of $\mathtt{D}$ is to distinguish in the most accurate way data coming from the real versus the synthetic distribution, while $\mathtt{G}$ tries to fool $\mathtt{D}$ by generating synthetic data that looks more and more like real.

Following the more general formulation introduced in~\cite{Nagarajan2017}, the GAN learning problem entails finding the minimax with respect to the pair $(\mathtt{G}, \mathtt{D})$ (i.e., the Nash equilibrium), of the value function defined as
\begin{equation}
  V(\mathtt{G},\mathtt{D}) = E_{p_{\mathcal{D}}(x)}[f(\mathtt{D}(x))] + E_{p(z)}[f(-\mathtt{D}(\mathtt{G}(z)))] \; ,
  \label{eq-gan}
\end{equation}
where $E[\cdot]$ denotes expectation, and $f :\mathbb{R} \rightarrow \mathbb{R}$ is a concave function. By setting $f(t) = -\log(1+\exp(-t))$ we obtain the original GAN formulation~\cite{Goodfellow2014}; instead, if $f(t) = t$ we obtain the Wasserstein GAN~\cite{Arjovsky2017}.

\section{Adversarial Latent Autoencoders}

\begin{figure}[t!]
\begin{center}
\includegraphics[width=1.0\linewidth]{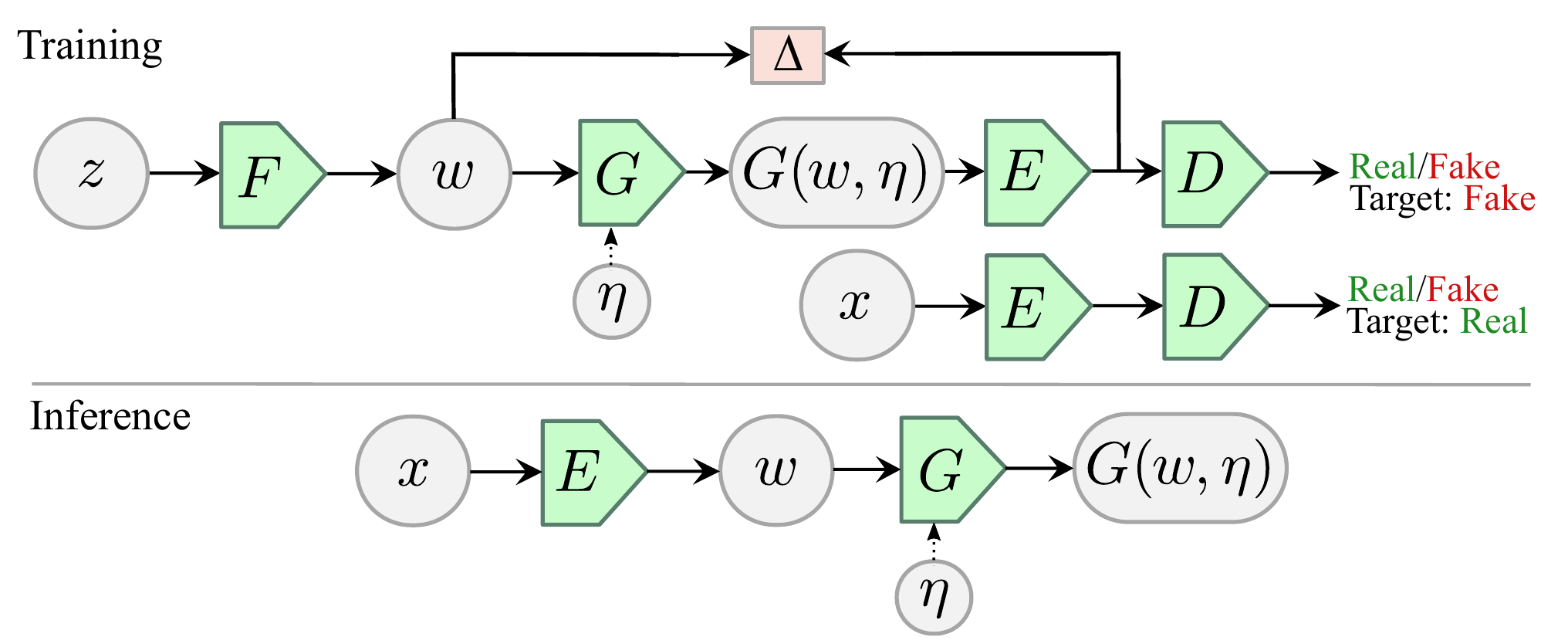}
\end{center}
\vspace{-4mm}
   \caption{\textbf{ALAE Architecture.} Architecture of an Adversarial Latent Autoencoder.}
\vspace{-3mm}
\label{fig-alae-architecture}
\end{figure}

We introduce a novel autoencoder architecture by modifying the original GAN paradigm. 
We begin by decomposing the generator $\mathtt{G}$ and the discriminator $\mathtt{D}$ in two networks: $F$, $G$, and $E$, $D$, respectively. This means that
\begin{equation}
  \mathtt{G} = G \circ F \; , \quad \mathrm{and} \quad \mathtt{D} = D \circ E \; ,
\end{equation}
see Figure~\ref{fig-alae-architecture}. In addition, we assume that the \emph{latent} spaces at the interface between $F$ and $G$, and between $E$ and $D$ are the same, and we indicate them as $\mathcal{W}$. In the most general case we assume that $F$ is a deterministic map, whereas we allow $E$ and $G$ to be stochastic. In particular, we assume that $G$ might optionally depend on an independent noisy input $\eta$, with a known fixed distribution $p_{\eta}(\eta)$. We indicate with $G(w,\eta)$ this more general stochastic generator.

Under the above conditions we now consider the distributions at the output of every network. The network $F$ simply maps $p(z)$ onto $q_F(w)$. At the output of $G$ the distribution can be written as
\begin{equation}
  q(x) = \int_w \int_{\eta} q_G(x|w,\eta) q_F(w) p_{\eta}(\eta) \; \mathrm{d} \eta \, \mathrm{d} w \; ,
\end{equation}
where $q_G(x|w,\eta)$ is the conditional distribution representing $G$. Similarly, for the output of $E$ the distribution becomes 
\begin{equation}
  q_E(w) = \int_x q_E(w|x) q(x) \mathrm{d} x \; ,
  \label{eq-qe}
\end{equation}
where $q_E(w|x)$ is the conditional distribution representing $E$. In~\eqref{eq-qe} if we replace $q(x)$ with $p_{\mathcal{D}}(x)$ we obtain the distribution $q_{E,\mathcal{D}}(w)$, which describes the output of $E$ when the real data distribution is its input.

Since optimizing~\eqref{eq-gan} leads toward  the synthetic distribution matching the real one, i.e., $q(x) = p_{\mathcal{D}}(x)$, it is obvious from~\eqref{eq-qe} that doing so also leads toward having $q_E(w) = q_{E,\mathcal{D}}(w)$. In addition to that, we propose to ensure that the distribution of the output of $E$ be the same as the distribution at the input of $G$. This means that we set up an additional goal, which requires that
\begin{equation}
  q_F(w) = q_E(w) \; .
  \label{eq-alae-condition}
\end{equation}
In this way we could interpret the pair of networks $(G,E)$ as a \emph{generator-encoder} network that autoencodes the \emph{latent space} $\mathcal{W}$.

If we indicate with $\Delta(p \|q)$ a measure of discrepancy between two distributions $p$ and $q$, we propose to achieve the goal~\eqref{eq-alae-condition} via regularizing the GAN loss~\eqref{eq-gan} by alternating the following two optimizations
\begin{eqnarray}
&  \min_{F,G} \max_{E,D} V(G \circ F, D \circ E) \label{eq-alae1}   \\
&  \min_{E,G}\Delta (F \| E\circ G \circ F)
\label{eq-alae2}  
\end{eqnarray}
where the left and right arguments of $\Delta$ indicate the distributions generated by the networks mapping $p(z)$, which correspond to  $q_F(w)$ and $q_E(w)$, respectively. We refer to a network optimized according to~\eqref{eq-alae1}~\eqref{eq-alae2} as an \emph{Adversarial Latent Autoencoder (ALAE)}. The building blocks of an ALAE architecture are depicted in Figure~\ref{fig-alae-architecture}.

\subsection{Relation with other autoencoders}
\label{sec-relation-to-ae}
  
\textbf{Data distribution.}
In architectures composed by an \emph{encoder} network and a \emph{generator} network, the task of the encoder is to map input data onto a space characterized by a \emph{latent distribution}, whereas the generator is tasked to map latent codes onto a space described by a \emph{data distribution}. Different strategies are used to learn the data distribution. For instance, some approaches impose a similarity criterion on the output of the generator~\cite{Kingma2014,Rezende2014, Makhzani2015, tomczak2017vae}, or even learn a similarity metric~\cite{Larsen2016}. Other techniques instead, set up an adversarial game to ensure the generator output matches the training data distribution~\cite{Donahue2016,Dumoulin2016,Srivastava2017,Ulyanov2018, Huang2018}. This latter approach is what we use for ALAE.
\begin{table}[t!]
  \centering
  \resizebox{0.8\columnwidth}{!}{%
    \footnotesize
  \begin{tabular}{ |p{1.5cm}||p{1.2cm}|p{2.2cm}|p{1.68cm}|  }
 \hline
Autoencoder& (a) Data & (b) Latent & (c) Reciprocity \\
 & Distribution & Distribution & Space\\
 \hline
 VAE~\cite{Kingma2014,Rezende2014}   & similarity  & imposed/divergence & data \\
 AAE~\cite{Makhzani2015} &  similarity  & imposed/adversarial  & data \\
 VAE/GAN~\cite{Larsen2016} & similarity & imposed/divergence & data \\
 VampPrior~\cite{tomczak2017vae} & similarity & learned/divergence & data \\
 BiGAN~\cite{Donahue2016}  & adversarial & imposed/adversarial & adversarial \\
 ALI~\cite{Dumoulin2016} &  adversarial & imposed/adversarial & adversarial \\
 VEEGAN~\cite{Srivastava2017} & adversarial & imposed/divergence & latent \\
 AGE~\cite{Ulyanov2018} & adversarial  & imposed/adversarial & latent \\
 IntroVAE~\cite{Huang2018} & adversarial & imposed/adversarial & data \\
 ALAE (ours) & adversarial & learned/divergence & latent \\
 \hline
\end{tabular}}
\caption{Autoencoder criteria used: (a) for matching the real to the synthetic data distribution; (b) for setting/learning the latent distribution; (c) for which space reciprocity is achieved.}
\label{tab-ae}
\vspace{-3mm}
\end{table}

\textbf{Latent distribution.}
For the latent space instead, the common practice is to set a desired target latent distribution, and then the encoder is trained to match it either by minimizing a divergence type of similarity~\cite{Kingma2014,Rezende2014,Larsen2016,Srivastava2017, tomczak2017vae}, or by setting up an adversarial game~\cite{Makhzani2015,Donahue2016,Dumoulin2016,Ulyanov2018,Huang2018}. Here is where ALAE takes a fundamentally different approach. Indeed, we do not impose the latent distribution, i.e., $q_E(w)$, to match a target distribution. The only condition we set, is given by~\eqref{eq-alae-condition}. In other words, we do not want $F$ to be the identity map, and are very much interested in letting the learning process decide what $F$ should be.


\textbf{Reciprocity.} 
Another aspect of autoecoders is whether and how they achieve reciprocity. This property relates to the ability of the architecture to reconstruct a data sample $x$ from its code $w$, and viceversa. Clearly, this requires that $x = G(E(x))$, or equivalently that $w = E(G(w))$. In the first case, the network must contain a reconstruction term that operates in the data space. In the latter one, the term operates in the latent space. While most approaches follow the first strategy~\cite{Kingma2014,Rezende2014,Makhzani2015,Larsen2016,Huang2018,tomczak2017vae}, there are some that implement the second~\cite{Srivastava2017,Ulyanov2018}, including ALAE. Indeed, this can be achieved by choosing the divergence in~\eqref{eq-alae2} to be the expected coding reconstruction error, as follows
\begin{equation}
  \Delta (F \| E \circ G \circ F ) = E_{p(z)} \left[ \| F(z) - E \circ G \circ F (z) \|_2^2\right]
\end{equation}
Imposing reciprocity in the latent space gives the significant advantage that simple $\ell_2$, $\ell_1$ or other norms can be used effectively, regardless of whether they would be inappropriate for the data space. For instance, it is well known that element-wise $\ell_2$ norm on image pixel differences does not reflect human visual perception. On the other hand, when used in latent space its meaning is different. For instance, an image translation by a pixel could lead to a large $\ell_2$ discrepancy in image space, while in latent space its representation would hardly change at all. Ultimately, using $\ell_2$ in image space has been regarded as one of the reasons why autoencoders have not been as successful as GANs in reconstructing/generating sharp images~\cite{Larsen2016}. Another way to address the same issue is by imposing reciprocity adversarially, as it was shown in~\cite{Donahue2016, Dumoulin2016}. Table~\ref{tab-ae} reports a summary of the main characteristics of most of the recent generator-encoder architectures.


%% file: main_stylealae.tex
\section{StyleALAE}
\label{section:style}
\begin{figure}[t!]
\begin{center}
\includegraphics[width=1.0\linewidth]{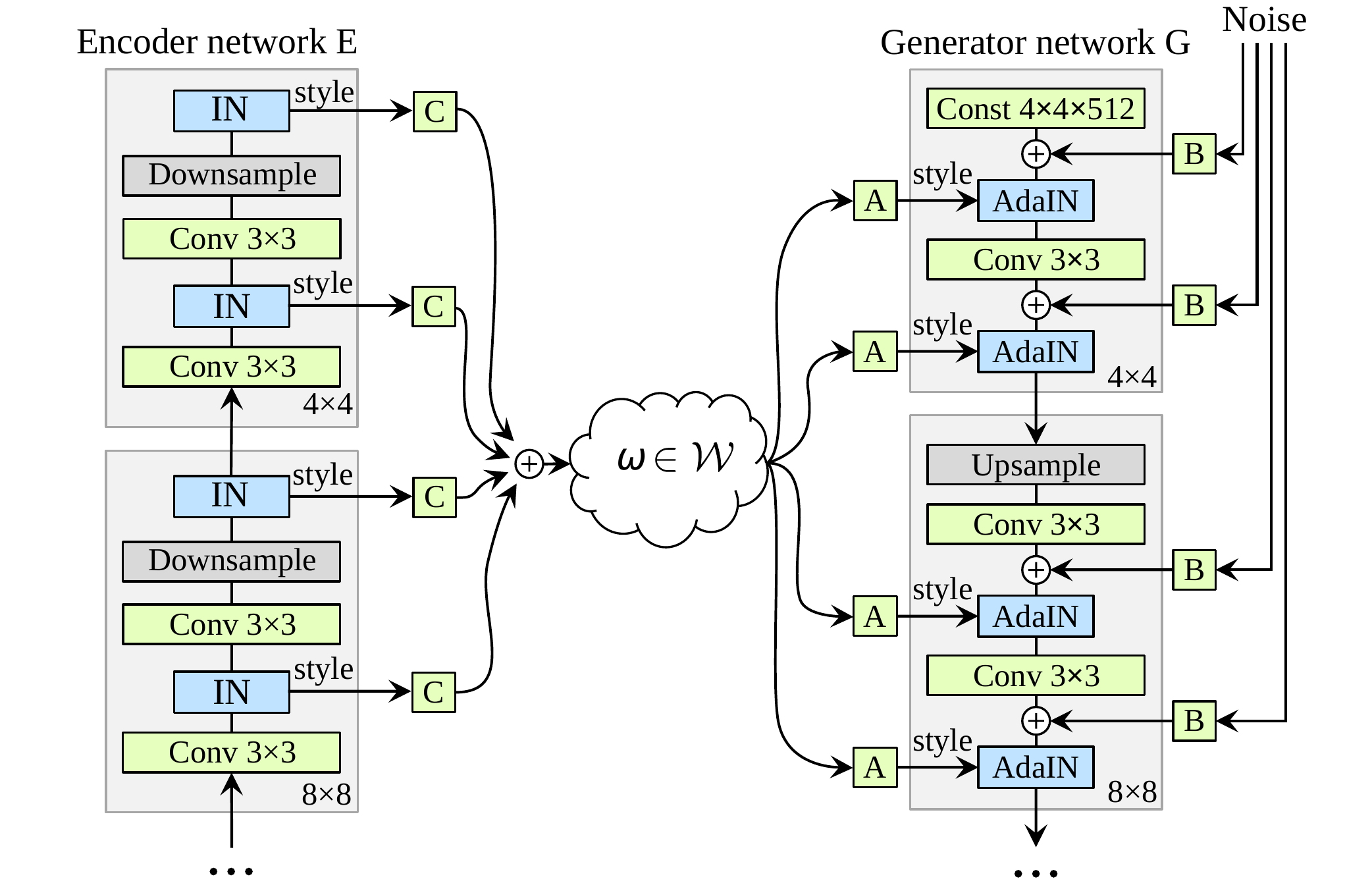}
\end{center}
\vspace{-4mm}
   \caption{\textbf{StyleALAE Architecture.} The StyleALAE encoder has Instance Normalization (IN) layers to extract multiscale style information that is combined into a latent code $w$ via a learnable multilinear map.}
\vspace{-3mm}
\label{fig-alae-style-encoder}
\end{figure}

We use ALAE to build an autoencoder that uses a StyleGAN based generator. For this we make our latent space $\mathcal{W}$ play the same role as the intermediate latent space in~\cite{Karras2019}. Therefore, our $G$ network becomes the part of StyleGAN depicted on the right side of Figure~\ref{fig-alae-style-encoder}. The left side is a novel architecture that we designed to be the encoder $E$.

Since at every layer, $G$ is driven by a style input, we design $E$ symmetrically, so that from a corresponding layer we extract style information. We do so by inserting Instance Normalization (IN) layers~\cite{huang2017arbitrary}, which provide instance averages and standard deviations for every channel. Specifically, if $y^E_i$ is the output of the $i$-th layer of $E$, the IN module extracts the statistics $\mu(y^E_i)$ and $\sigma(y^E_i)$ representing the style at that level. The IN module also provides as output the normalized version of the input, which continues down the pipeline with no more style information from that level. Given the information flow between $E$ and $G$, the architecture is effectively mimicking a multiscale style transfer from $E$ to $G$, with the difference that there is not an extra input image that provides the content~\cite{huang2017arbitrary,huang2018multimodal}.

The set of styles that are inputs to the Adaptive Instance Normalization (AdaIN) layers~\cite{huang2017arbitrary} in $G$ are related linearly to the latent variable $w$. Therefore, we propose to combine the styles output by the encoder, and to map them onto the latent space, via the following multilinear map
\begin{equation}
w = \sum_{i=1}^{N}C_i\begin{bmatrix}
\mu(y^E_i) \\
\sigma(y^E_i)
\end{bmatrix}
  \label{eq-encoder}
\end{equation} 
where the $C_i$'s are learnable parameters, and $N$ is the number of layers.

Similarly to~\cite{karras2017progressive, Karras2019} we use progressive growing. We start from low-resolution images ($4\times4$ pixels) and progressively increase the resolution by smoothly blending in new blocks to $E$ and $G$. For the $F$ and $D$ networks we implement them using MLPs. The $\mathcal{Z}$ and $\mathcal{W}$ spaces, and all layers of $F$ and $D$ have the same dimensionality in all our experiments.  Moreover, for StyleALAE we follow~\cite{Karras2019}, and chose $F$ to have 8 layers, and we set $D$ to have 3 layers.
\begin{algorithm}[b!]
\caption{ALAE Training}
\label{alg:training}
\footnotesize
\begin{algorithmic}[1]
\State \(\theta_F, \theta_G, \theta_E, \theta_D \gets\) Initialize~ network~ parameters
\While{not converged}
\State Step I. Update $E$, and $D$ 
\State \(x \gets\) Random mini-batch from dataset
\State \(z \gets\) Samples from prior \(\mathcal{N}(0,I)\)
\State \(L^{E,D}_{adv} \gets \SP(D \circ E \circ G \circ F (z))) + \SP(-D \circ E (x)) + \frac{\gamma}{2} \E_{ p_{\mathcal D}(x)}\left[\|\nabla D \circ E (x)\|^2\right] \)
\State \( \theta_E, \theta_D \gets \textsc{Adam}(\nabla_{\theta_D, \theta_E} L^{E,D}_{adv}, \theta_D, \theta_E, \alpha, \beta_1, \beta_2)\)

\State Step II. Update $F,$ and $G$ 
\State \(z \gets\) Samples from prior \(\mathcal{N}(0,I)\)
\State \(L^{F,G}_{adv} \gets \SP(-D \circ E \circ G \circ F(z))) \)
\State \( \theta_F, \theta_G \gets \textsc{Adam}(\nabla_{\theta_F, \theta_G} L^{F,G}_{adv}, \theta_F, \theta_G, \alpha, \beta_1, \beta_2)\)

\State Step III. Update $E$, and $G$
\State \(z \gets\) Samples from prior \(\mathcal{N}(0,I)\)
\State \(L^{E,G}_{error} \gets \| F(z) - E \circ G \circ F (z) \|_2^2\)

\State \(\theta_E, \theta_G \gets \textsc{Adam}(\nabla_{\theta_E, \theta_G} L^{E,G}_{error}, \theta_E, \theta_G, \alpha, \beta_1, \beta_2)\)
\EndWhile
\end{algorithmic}
\end{algorithm}

\section{Implementation}

\textbf{Adversarial losses and regularization.}
We use a non-saturating loss~\cite{Goodfellow2014,mescheder2018training}, which in~\eqref{eq-gan} we introduce by setting $f(\cdot)$ to be a SoftPlus function~\cite{glorot2011deep}. This is a smooth version of the rectifier activation function, defined as $f(t) = \SP (t) = \log(1+\exp(t))$. In addition, we use gradient regularization techniques~\cite{drucker1992improving,mescheder2018training, ross2018improving}. We utilize $R_1$~\cite{roth2017stabilizing, mescheder2018training}, a zero-centered gradient penalty term which acts only on real data, and is defined as $\frac{\gamma}{2} \E_{ p_{\mathcal D}(x)}\left[\|\nabla D \circ E (x)\|^2\right]$, where the gradient is taken with respect to the parameters $\theta_E$ and $\theta_D$ of the networks $E$ and $D$, respectively.


\textbf{Training.}
In order to optimizate~\eqref{eq-alae1}~\eqref{eq-alae2} we use alternating updates. One iteration is composed of three updating steps: two for~\eqref{eq-alae1} and one for \eqref{eq-alae2}. Step I updates the discriminator (i.e., networks $E$ and $D$). Step II updates the generator (i.e., networks $F$ and $G$). Step III updates the latent space autoencoder (i.e., networks $G$ and $E$). The procedural details are summarized in Algorithm~\ref{alg:training}. For updating the weights we use the Adam optimizer~\cite{kingma2014adam} with $\beta_1 = 0.0$ and $\beta_2 = 0.99$, coupled with the learning rate equalization technique~\cite{karras2017progressive} described below. For non-growing architectures (i.e., MLPs) we use a learning rate of $0.002$, and batch size of 128. For growing architectures (i.e., StyleALAE) learning rate and batch size depend on the resolution.



%% file: main_experiments1.tex
\section{Experiments}
Code and uncompressed images are available at \url{https://github.com/podgorskiy/ALAE}.
\begin{figure}[t!]
\centering
\begingroup
\renewcommand{\arraystretch}{1.7}
\begin{tabular}{rc@{}c@{}}
Input &
\includegraphics[width=0.79\linewidth]{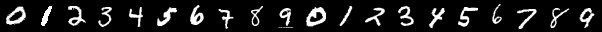} \\
\midrule
BiGAN &
\includegraphics[width=0.79\linewidth]{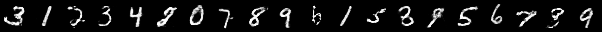} \\
ALAE &
\includegraphics[width=0.79\linewidth]{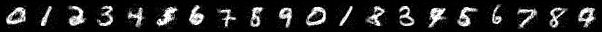} \\
\end{tabular}
\endgroup
\vspace{-3mm}
\caption{\textbf{MNIST reconstruction.} Reconstructions of the permutation-invariant MNIST. Top row: real images. Middle row: BiGAN reconstructions. Bottom row: ALAE reconstructions. The same MLP architecture is used in both methods.
}
\vspace{-3mm}
\label{f-mnist}
\end{figure}
\begin{figure}[b!]
\centering
\begin{tabular}{rc@{}c@{}}
$\mathcal{Z}$ space &
\includegraphics[width=0.74\linewidth]{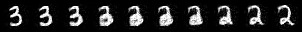} \\
\midrule
$\mathcal{W}$ space &
\includegraphics[width=0.74\linewidth]{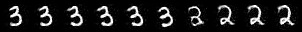}
\end{tabular}
\vspace{-3mm}
\caption{\textbf{MNIST traversal.} Reconstructions of the interpolations in the $\mathcal{Z}$ space, and the $\mathcal{W}$ space, between the same digits. The latter transition appears to be smoother.}
\label{fig-mnist-traversal}
\vspace{-3mm}
\end{figure}

\subsection{Representation learning with MLP}

We train ALAE with MNIST~\cite{lecun1998gradient}, and then use the feature representation for classification, reconstruction, and analyzing disentanglement. We use the permutation-invariant setting, where each $28 \times 28$ MNIST image is treated as a $784$D vector without spatial structure, which requires to use a MLP instead of a CNN. We follow~\cite{donahue2016adversarial} and use a three layer MLP with a latent space size of $50$D. Both networks, $E$ and $G$ have two hidden layers with 1024 units each. In~\cite{donahue2016adversarial} the features used are the activations of the layer before the last of the encoder, which are $1024$D vectors. We refer to those as \emph{long features}. We also use, as features, the $50$D vectors taken from the latent space, $\mathcal{W}$. We refer to those as \emph{short features}.

MNIST has an official split into training and testing sets of sizes 60000 and 10000 respectively. We refer to it as \emph{different writers} (DW) setting since the human writers of the digits for the training set are different from those who wrote the testing digits. We consider also a \emph{same writers} (SW) setting, which uses only the official training split by further splitting it in two parts: a train split of size 50000 and a test split of size 10000, while the official testing split is ignored. In SW  the pools of writers in the train and test splits overlap, whereas in DW they do not. This makes SW an easier setting than DW.

\begin{table}[t]
  \caption{\textbf{MNIST classification.} Classification accuracy (\%) on the permutation-invariant MNIST~\cite{lecun1998gradient} using 1NN and linear SVM, with \emph{same writers} (SW) and \emph{different writers} (DW) settings, and short features (sf) vs. long features (lf), indicated as sf/lf.  }
  \centering
  \resizebox{\columnwidth}{!}{%
    \begin{tabular}{lcccc}
      \toprule
      & \bf{1NN SW} & \bf{Linear SVM SW}  & \bf{1NN DW}  & \bf{Linear SVM DW} \\
      \midrule
      AE($\ell_1$)     & \underline{97.15}/\underline{97.43} & 88.71/97.27 & \underline{96.84}/96.80 & 89.78/97.72 \\
      AE($\ell_2$)     & {\bf 97.52}/97.37 & 88.78/97.23 & {\bf 97.05}/96.77 & 89.78/97.72 \\
      LR         & 92.79/97.28 & 89.74/97.56 & 91.90/96.69 & 90.03/\underline{97.80} \\
      JLR        & 92.54/97.02 & 89.23/97.19 & 91.97/96.45 & 90.82/97.62 \\ 
      BiGAN~\cite{donahue2016adversarial}      & 95.83/97.14 & \underline{90.52}/\underline{97.59} & 95.38/\underline{96.81} & \underline{91.34}/97.74 \\
      ALAE (ours) & 93.79/{\bf 97.61} & {\bf 93.47}/{\bf 98.20} & 94.59/{\bf 97.47} & {\bf 94.23}/{\bf 98.64} \\         
      \bottomrule
    \end{tabular}}
  \label{t-mnist}
  \vspace{-4mm}
\end{table}

\textbf{Results.} We report the accuracy with the 1NN classifier as in~\cite{donahue2016adversarial}, and extend those results by reporting also the accuracy with the linear SVM, because it allows a more direct analysis of disentanglement. Indeed, we recall that a disentangled representation~\cite{schmidhuber1992learning,ridgeway2016survey,achille2018emergence} refers to a space consisting of linear subspaces, each of which is responsible for one factor of variation. Therefore, a linear classifier based on a disentangled feature space should lead to better performance compared to one working on an entangled space. Table~\ref{t-mnist} summarizes the average accuracy over five trials for ALAE, BiGAN, as well as the following baselines proposed in~\cite{donahue2016adversarial}: Latent Regressor (LR), Joint Latent Regressor (JLR), Autoencoders trained to minimize the $\ell_2$ (AE($\ell_2$)) or the $\ell_1$ (AE($\ell_1$)) reconstruction error. 

The most significant result of Table~\ref{t-mnist} is drawn by comparing the 1NN with the corresponding linear SVM columns. Since 1NN does not presume disentanglement in order to be effective, but linear SVM does, larger performance drops signal stronger entanglement. ALAE is the approach that remains more stable when switching from 1NN to linear SVM, suggesting a greater disentanglement of the space. This is true especially for short features, whereas for long features this effect fades away because linear separability grows. 

We also note that ALAE does not always provide the best accuracy, and the baseline AE (especially AE($\ell_2$)) does well with 1NN, and more so with short features. This might be explained by the baseline AE learning a representation that is closer to a discriminative one. Other approaches instead focus more on learning representations for drawing synthetic random samples, which are likely richer, but less discriminative. This effect also fades for longer features.

Another observation is about SW vs. DW. 1NN generalizes less effectively for DW, as expected, but linear SVM provides a small improvement. This is unclear, but we speculate that DW might have fewer writers in the test set, and potentially slightly less challenging.

Figure~\ref{f-mnist} shows qualitative reconstruction results. It can be seen that BiGAN reconstructions are subject to semantic label flipping much more often than ALAE. Finally, Figure~\ref{fig-mnist-traversal} shows two traversals: one obtained by interpolating in the $\mathcal{Z}$ space, and the other by interpolating in the $\mathcal{W}$ space. The second shows a smoother image space transition, suggesting a lesser degree of entanglement.

%% file: main_experiments2.tex
\begin{figure*}[t!]
\begin{center}
\includegraphics[width=1.0\linewidth]{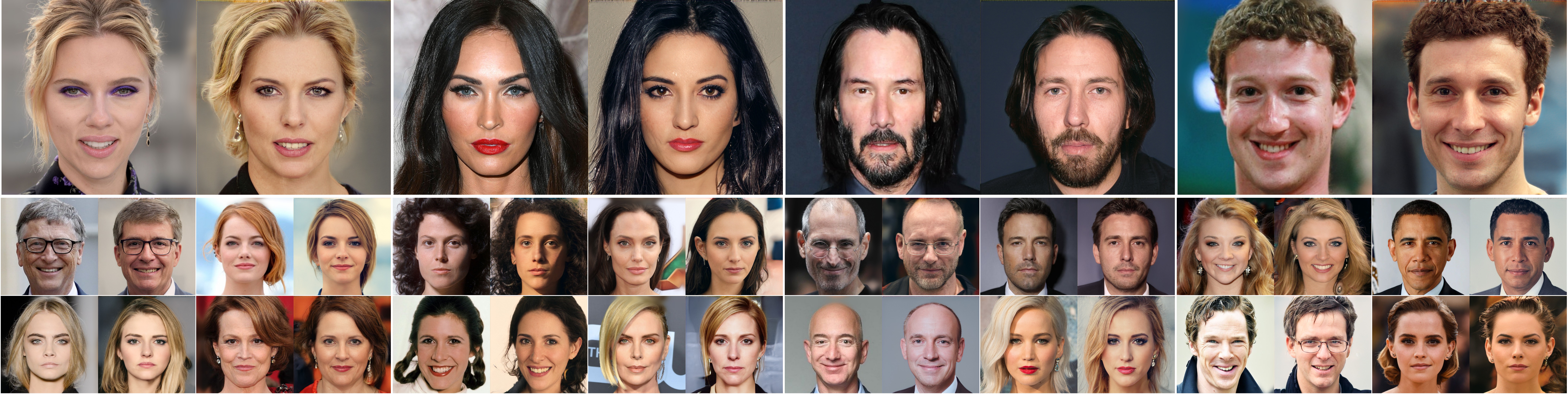}
\end{center}
\vspace{-5mm}
   \caption{\textbf{FFHQ reconstructions.} Reconstructions of unseen images with StyleALAE trained on FFHQ~\cite{Karras2019} at $1024 \times 1024$.}
\label{fig:recffhq}
\vspace{-4mm}
\end{figure*}
\begin{table}[t!]
\centering
\newcolumntype{x}{>{\centering\arraybackslash\hspace{0pt}}p{11.2mm}}
\resizebox{0.7\columnwidth}{!}{%
\footnotesize{
\begin{tabular}{|l@{\hspace{1mm}}|c|x|x|}
\hline
  \textbf{Method}                    & \textbf{FFHQ}     & \textbf{LSUN Bedroom}      \\\hline\hline
  StyleGAN~\cite{Karras2019}                             & 4.40              & 2.65             \\\hline %
  PGGAN~\cite{karras2017progressive} & -                 & 8.34             \\\hline %
  IntroVAE~\cite{Huang2018}          & -                 & 8.84             \\\hline %
  Pioneer~\cite{heljakka2018pioneer}                            & -                 & 18.39             \\\hline %
  Balanced Pioneer~\cite{heljakka2019towards}                   & -                 & 17.89             \\\hline %
  StyleALAE Generation                    & 13.09             & 17.13             \\\hline %
  StyleALAE Reconstruction                & 16.52             & 15.92   \\\hline %
\end{tabular}
}
}
\caption{\textbf{FID scores.} FID scores (lower is better) measured on FFHQ~\cite{Karras2019} and LSUN Bedroom~\cite{yu2015lsun}.
}
\label{tab:fid}
\vspace{-4mm}
\end{table}
\subsection{Learning style representations}

\textbf{FFHQ.} We evaluate StyleALAE with the FFHQ~\cite{Karras2019} dataset. It is very recent and consists of 70000 images of people faces aligned and cropped at resolution of $1024 \times 1024$. In contrast to~\cite{Karras2019}, we split FFHQ into a training set of 60000 images and a testing set of 10000 images. We do so in order to measure the reconstruction quality for which we need images that were not used during training.

We implemented our approach with PyTorch. Most of the experiments were conducted on a machine with 4$\times$ GPU Titan X, but for training the models at resolution $1024\times1024$ we used a server with 8$\times$ GPU Titan RTX. We trained StyleALAE for 147 epochs, 18 of which were spent at resolution $1024\times1024$. Starting from resolution $4\times 4$ we grew StyleALAE up to $1024\times1024$. When growing to a new resolution level we used $500$k training samples during the transition, and another $500$k samples for training stabilization. Once reached the maximum resolution of $1024\times1024$, we continued training for $1$M images. Thus, the total training time measured in images was $10$M. In contrast, the total training time for StyleGAN~\cite{Karras2019} was $25$M images, and $15$M of them were used at resolution $1024\times1024$. At the same resolution we trained StyleALAE with only $1$M images, so, 15 times less.

\begin{table}[t!]
\centering
\newcolumntype{x}{>{\centering\arraybackslash\hspace{0pt}}p{11.2mm}}
\resizebox{0.7\columnwidth}{!}{%
\footnotesize{
\begin{tabular}{|l@{\hspace{1mm}}c|x|x|}
\hline
  \multirow{2}{*}{\textbf{Method}}  &           & \multicolumn{2}{c|}{\textbf{Path length}} \\
                                    &           & \textbf{full}     & \textbf{end}      \\\hline\hline
  StyleGAN                          & $\mathcal{Z}$     & 412.0             & 415.3             \\\hline %
  StyleGAN no mixing                & $\mathcal{W}$     & 200.5             & 160.6             \\\hline %
  StyleGAN                          & $\mathcal{W}$     & 231.5             & 182.1             \\\hline %
  StyleALAE                              & $\mathcal{Z}$     & 300.5             & 292.0             \\\hline %
  StyleALAE                              & $\mathcal{W}$     & \textbf{134.5}    & \textbf{103.4}    \\\hline %
\end{tabular}
}
}
\caption{\textbf{PPL.} Perceptual path lengths on FFHQ measured in the $\mathcal{Z}$ and the $\mathcal{W}$ spaces (lower is better).
}
\label{tab:disentangle}
\vspace{-4mm}
\end{table}
\begin{table}[b!]
\vspace{-4mm}
  \centering
  \newcolumntype{x}{>{\centering\arraybackslash\hspace{0pt}}p{11.2mm}}
\resizebox{0.7\columnwidth}{!}{%
  \footnotesize{
\begin{tabular}{|l@{\hspace{1mm}}|c|x|x|x|x|}
\hline
    & \textbf{FID}  & \textbf{PPL full}  \\\hline\hline
  PGGAN~\cite{karras2017progressive}  & $\bf 8.03 \bf$ & $229.2$  \\\hline %
  GLOW~\cite{Kingma2018}  & $68.93$ & $219.6$  \\\hline %
  PIONEER~\cite{heljakka2018pioneer}  & $39.17$ & 155.2 \\\hline %
  Balanced PIONEER~\cite{heljakka2019towards} &  $25.25$ & $146.2$ \\\hline %
  StyleALAE (ours) & 19.21 & $\bf 33.29$ \\\hline %
\end{tabular}}
}
  \caption{Comparison of FID and PPL scores for CelebA-HQ images at $256{\times}256$ (lower is better). FID is based on 50,000 generated samples compared to training samples.}
    \label{t-celeba}
\end{table}

Table~\ref{tab:fid} reports the FID score~\cite{heusel2017gans} for generations and reconstructions. Source images for reconstructions are from the test set and were not used during training. The scores of StyleALAE are higher, and we regard the large training time difference between StyleALAE and StyleGAN ($1$M vs $15$M) as the likely cause of the  discrepancy.
\begin{figure}[t!]
\begin{center}
\includegraphics[width=1.0\linewidth]{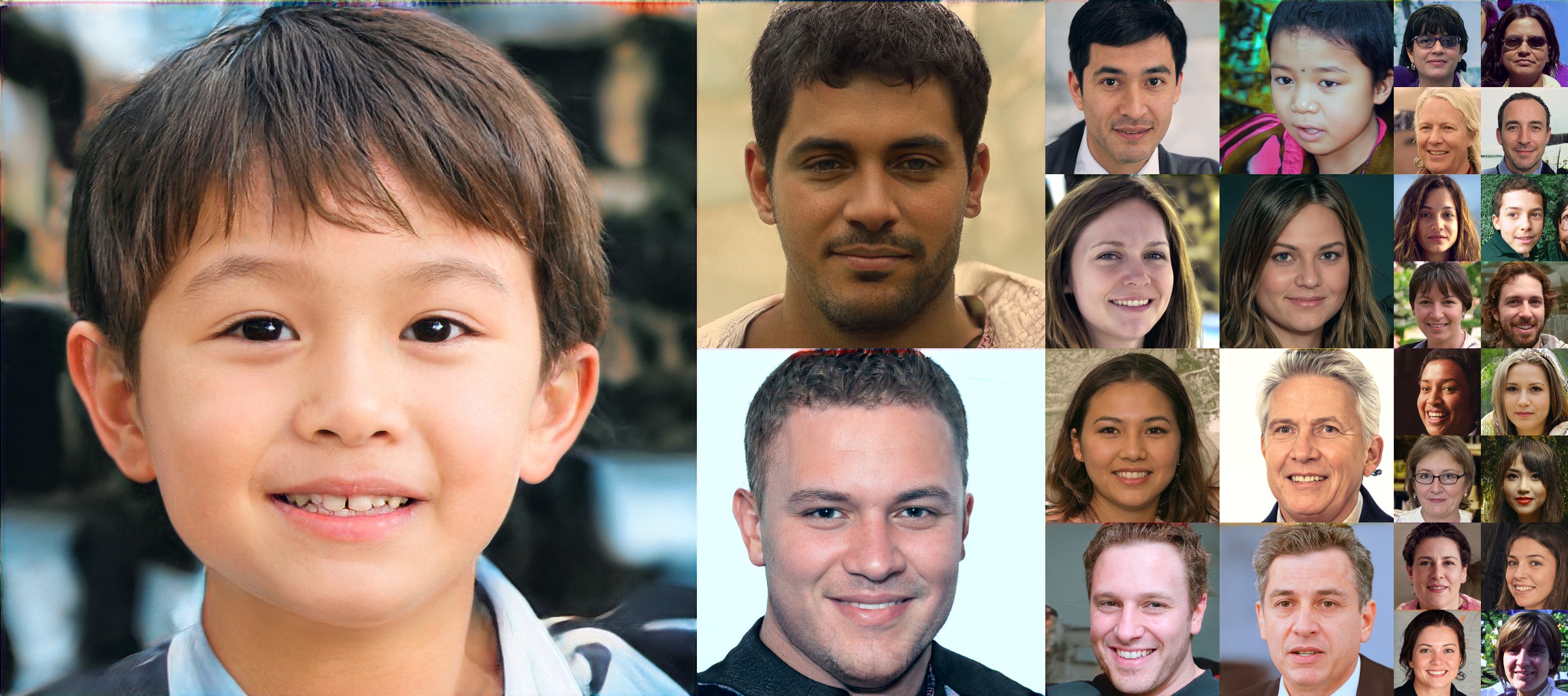}
\end{center}
\vspace{-4mm}
\caption{\textbf{FFHQ generations.} Generations with StyleALAE trained on FFHQ~\cite{Karras2019} at $1024 \times 1024$.}
\vspace{-1mm}
\label{fig:genffhq}
\end{figure}

Table~\ref{tab:disentangle} reports the perceptual path length (PPL)~\cite{Karras2019} of SyleALAE. This is a measurement of the degree of disentanglement of representations. We compute the values for representations in the $\mathcal{W}$ and the $\mathcal{Z}$ space, where StyleALAE is trained with style mixing in both cases. The StyleGAN score measured in $\mathcal{Z}$ corresponds to a traditional network, and in $\mathcal{W}$ for a style-based one. We see that the PPL drops from $\mathcal{Z}$ to $\mathcal{W}$, indicating that $\mathcal{W}$  is perceptually more linear than $\mathcal{Z}$, thus less entangled. Also, note that for our models the PPL is lower, despite the higher FID scores.

Figure~\ref{fig:genffhq} shows a random collection of generations obtained from
StyleALAE. Figure~\ref{fig:recffhq} instead shows a collection of reconstructions. In Figure~\ref{fig:stylemix} instead, we repeat the style mixing experiment in~\cite{Karras2019}, but with real images as sources and destinations for style combinations. We note that the original images are faces of celebrities that we downloaded from the internet. Therefore, they are not part of FFHQ, and come from a different distribution. Indeed, FFHQ is made of face images obtained from Flickr.com depicting non-celebrity people. Often the faces do not wear any makeup, neither have the images been altered (e.g., with Photoshop). Moreover, the imaging conditions of the FFHQ acquisitions are very different from typical photoshoot stages, where professional equipment is used. Despite this change of image statistics, we observe that StyleALAE works effectively on both reconstruction and mixing.

\begin{figure}[t!]
\begin{center}

  
\includegraphics[width=\linewidth]{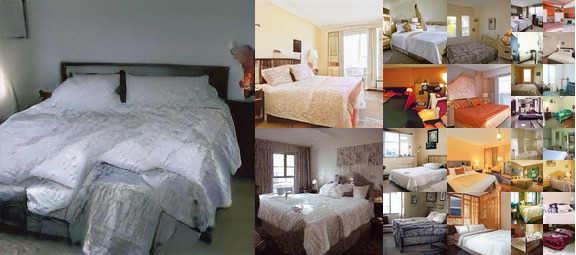}
\includegraphics[width=\linewidth]{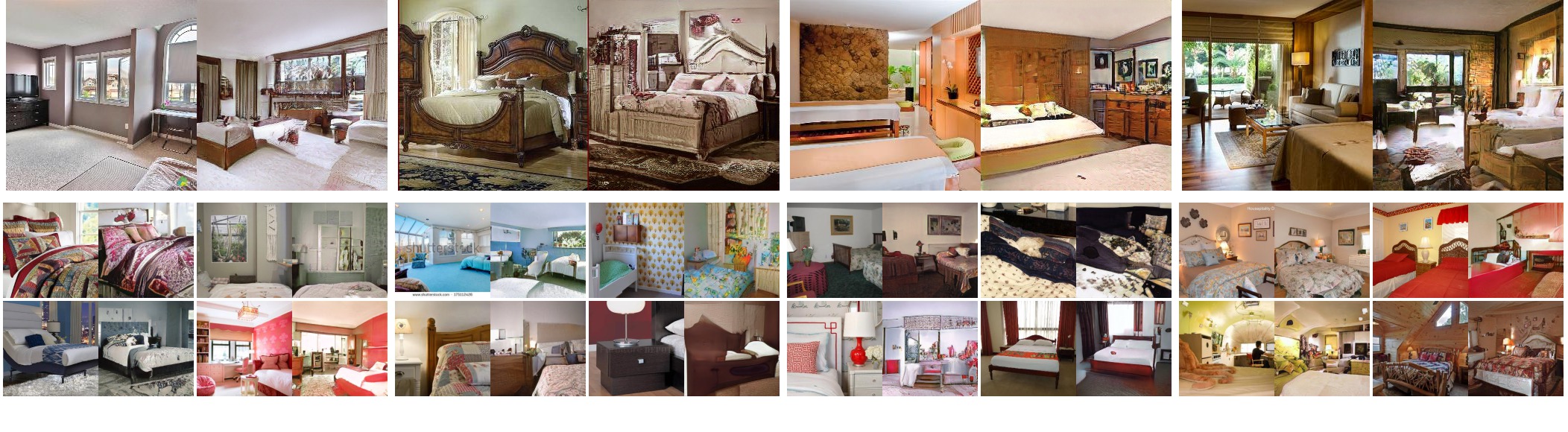}

\end{center}
\vspace{-5mm}
\caption{\textbf{LSUN generations and reconstructions.} Generations (first row), and reconstructions using StyleALAE trained on LSUN Bedroom~\cite{yu2015lsun} at resolution $256 \times 256$.}
\label{fig:genbed}
\vspace{-4mm}
\end{figure}

\textbf{LSUN.} We evaluated StyleALAE with LSUN Bedroom~\cite{yu2015lsun}.  Figure~\ref{fig:genbed} shows generations and reconstructions from unseen images during training. Table~\ref{tab:fid} reports the FID scores on the generations and the reconstructions.



\begin{figure}[t]
\begin{tabular}{c}
\includegraphics[width=0.23\linewidth]{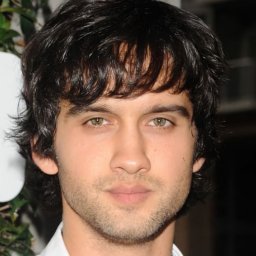}
\includegraphics[width=0.23\linewidth]{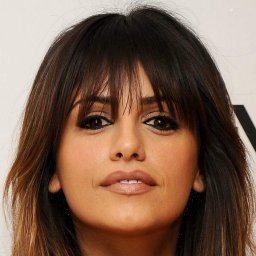}
\includegraphics[width=0.23\linewidth]{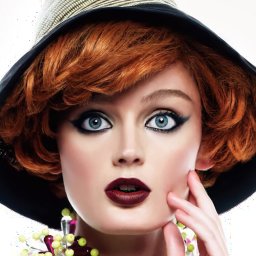}
\includegraphics[width=0.23\linewidth]{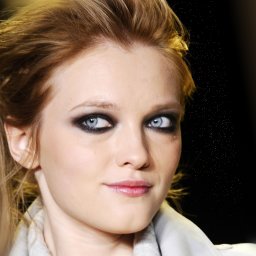}\\
\midrule
\includegraphics[width=0.23\linewidth]{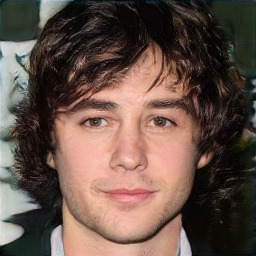}
\includegraphics[width=0.23\linewidth]{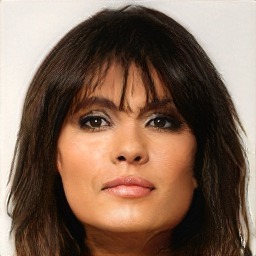}
\includegraphics[width=0.23\linewidth]{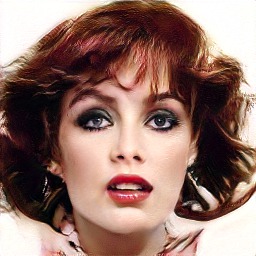}
\includegraphics[width=0.23\linewidth]{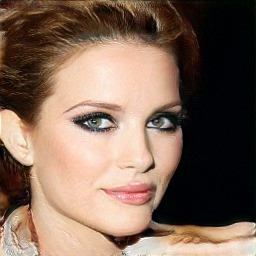}\\
\includegraphics[width=0.23\linewidth]{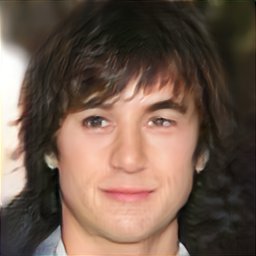}
\includegraphics[width=0.23\linewidth]{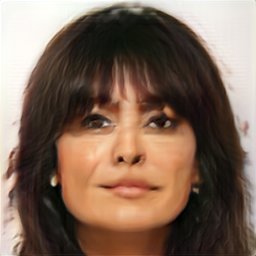}
\includegraphics[width=0.23\linewidth]{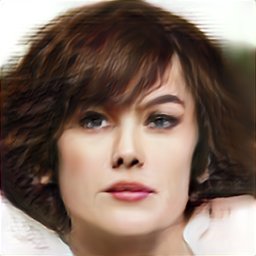}
\includegraphics[width=0.23\linewidth]{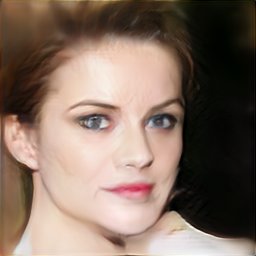}\\
\includegraphics[width=0.23\linewidth]{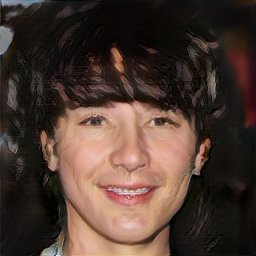}
\includegraphics[width=0.23\linewidth]{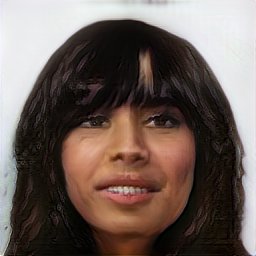}
\includegraphics[width=0.23\linewidth]{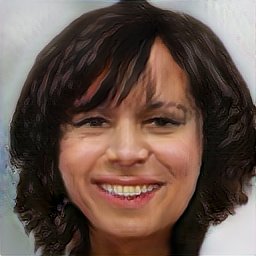}
\includegraphics[width=0.23\linewidth]{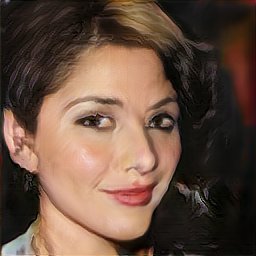}
\end{tabular}
\vspace{-3mm}
\caption{\textbf{CelebA-HQ reconstructions.} CelebA-HQ reconstructions of unseen samples at resolution $256 \times 256$. Top row: real images. Second row: StyleALAE. Third row: Balanced PIONEER~\cite{heljakka2019towards}. Last row: PIONEER~\cite{heljakka2018pioneer}. StyleALAE reconstructions look sharper and less distorted.}
\vspace{-4mm}
\label{f-ablation}
\end{figure}

\textbf{CelebA-HQ.} CelebA-HQ~\cite{karras2017progressive} is an improved subset of CelebA~\cite{liu2015deep} consisting of 30000 images at resolution $1024 \times 1024$. We follow~\cite{heljakka2018pioneer, heljakka2019towards, Kingma2018, karras2017progressive} and use CelebA-HQ downscaled to $256 \times 256$ with training/testing split of 27000/3000. Table~\ref{t-celeba} reports the FID and PPL scores, and Figure~\ref{f-ablation} compares StyleALE reconstructions of unseen faces with two other approaches.

\newcommand{\h}{0mm}
\newcommand{\hh}{0mm}
\newcommand{\hhh}{0mm}
\newcommand{\vv}{0mm}
\newcommand{\vvv}{0mm}
\newcommand{\ext}{jpg}  %
\newcommand{\extt}{jpg} %
\renewcommand{\h}{27.4mm}
\renewcommand{\hh}{2.5mm}
\renewcommand{\vv}{\vspace*{-0.35mm}}
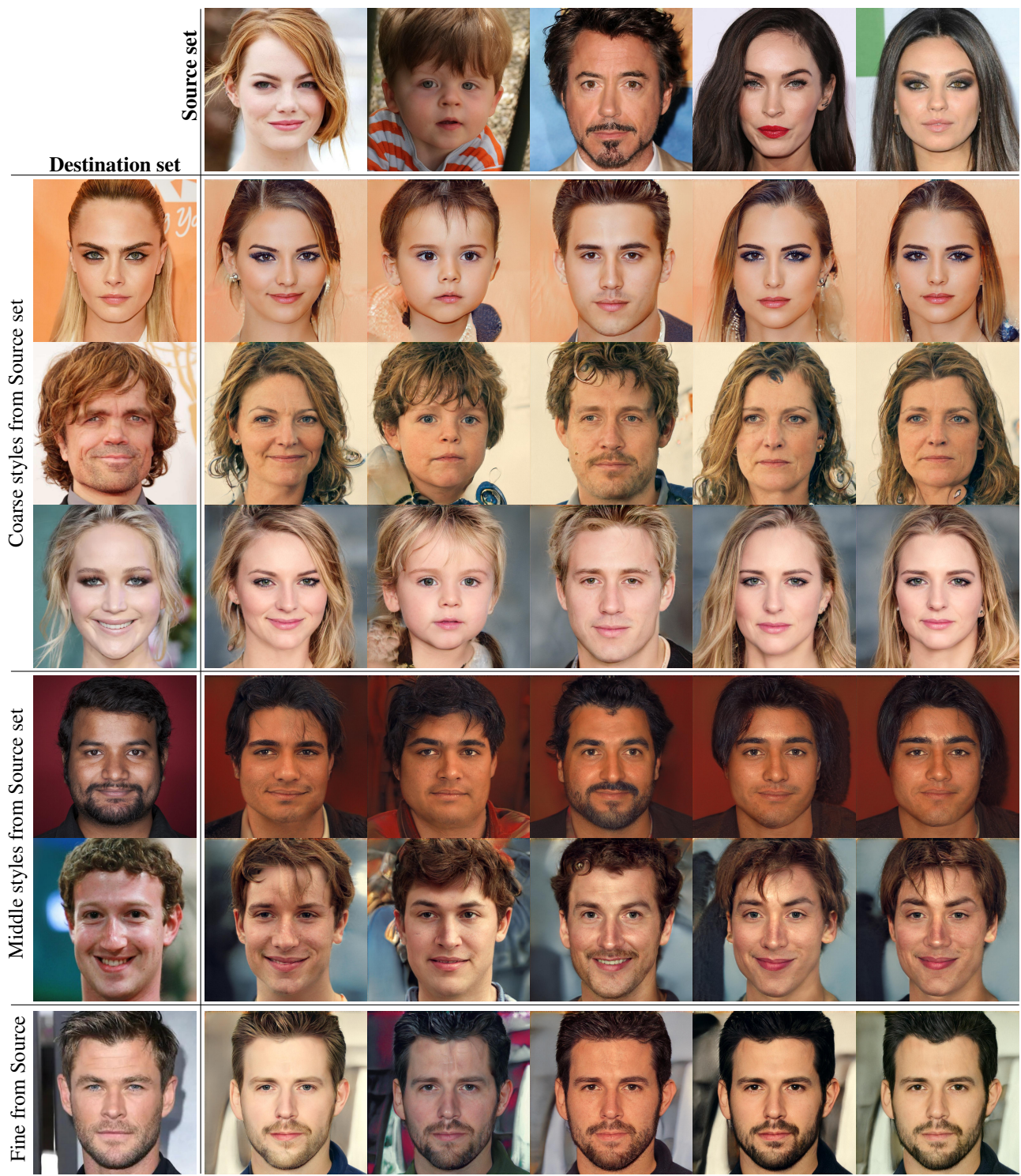
\begin{figure*}[t]
\centering
\begin{minipage}[c]{0.972\linewidth}
\makebox[\hh][c]{}\hfill%
\makebox[\h][c]{\textbf{\normalsize{\ Destination set}}}\hfill%
\rotatebox[origin=l]{90}{\makebox[0mm][l]{\hspace*{0.05\linewidth}\textbf{\normalsize{\raisebox{.5mm}[0mm][0mm]{Source set}}}}}\hfill%
\newcommand{\varA}{var639}\includegraphics[height=\h]{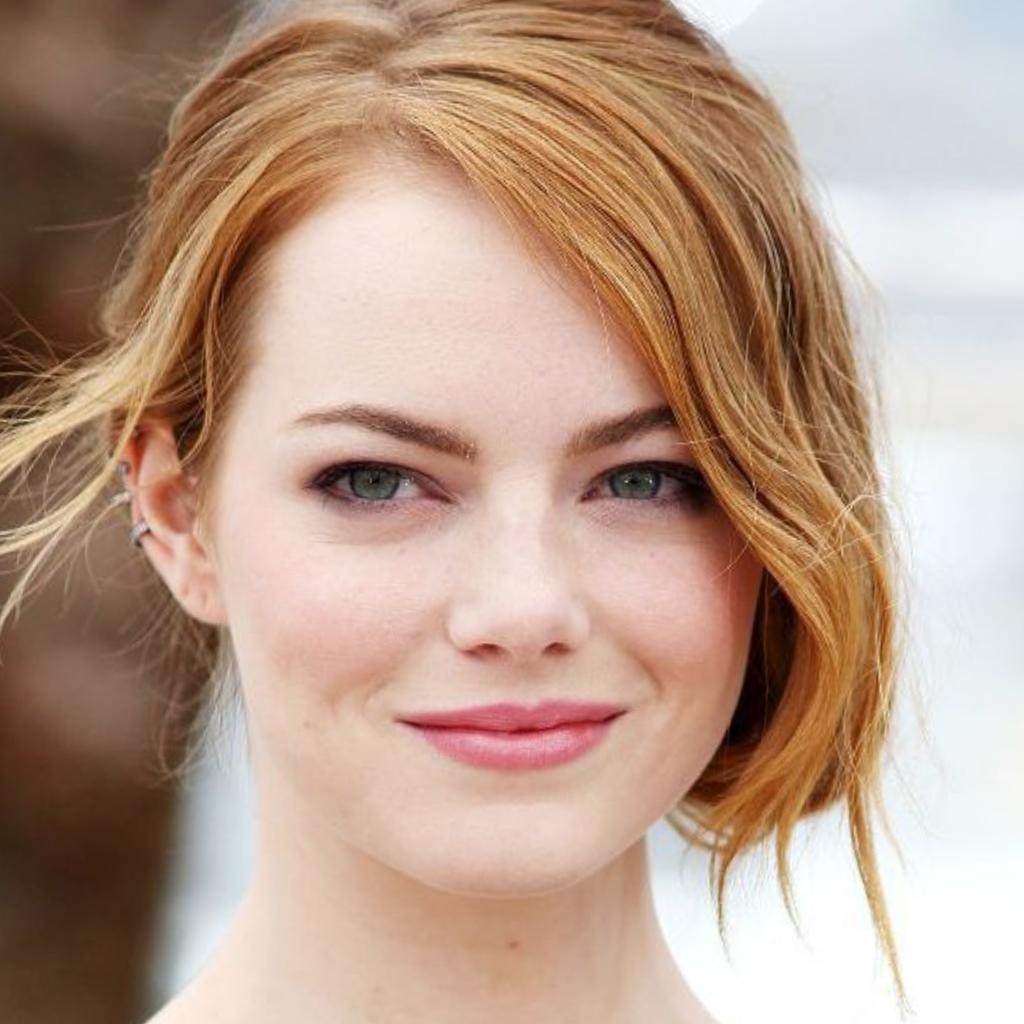}%
\newcommand{\varB}{var701}\includegraphics[height=\h]{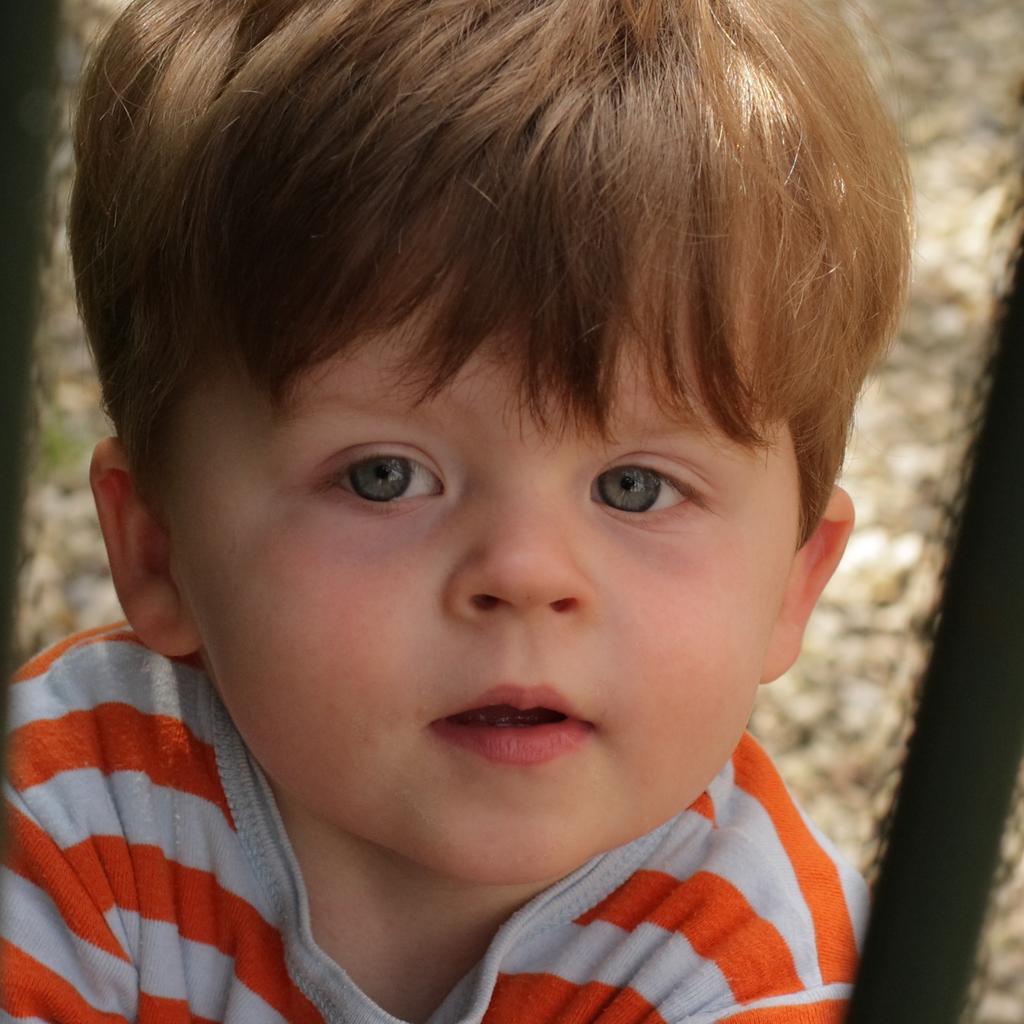}%
\newcommand{\varC}{var687}\includegraphics[height=\h]{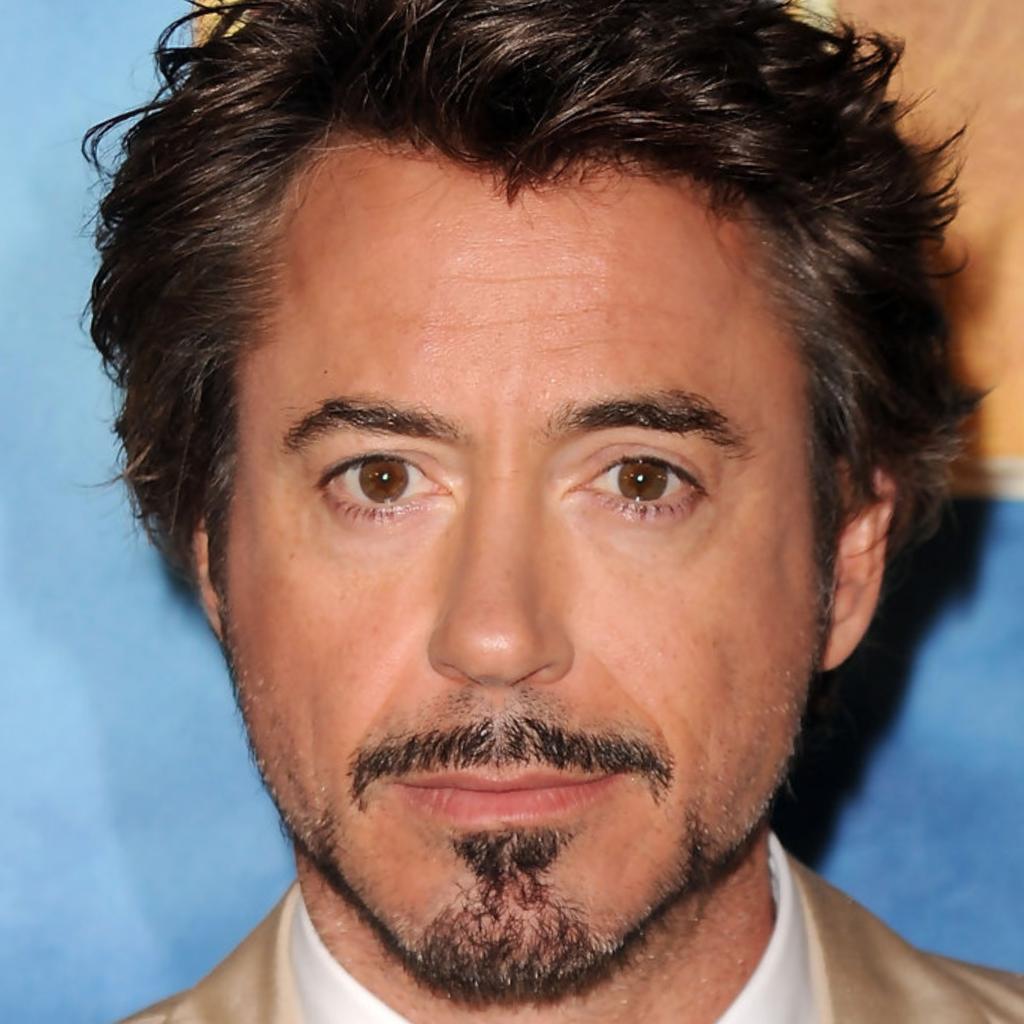}%
\newcommand{\varD}{var615}\includegraphics[height=\h]{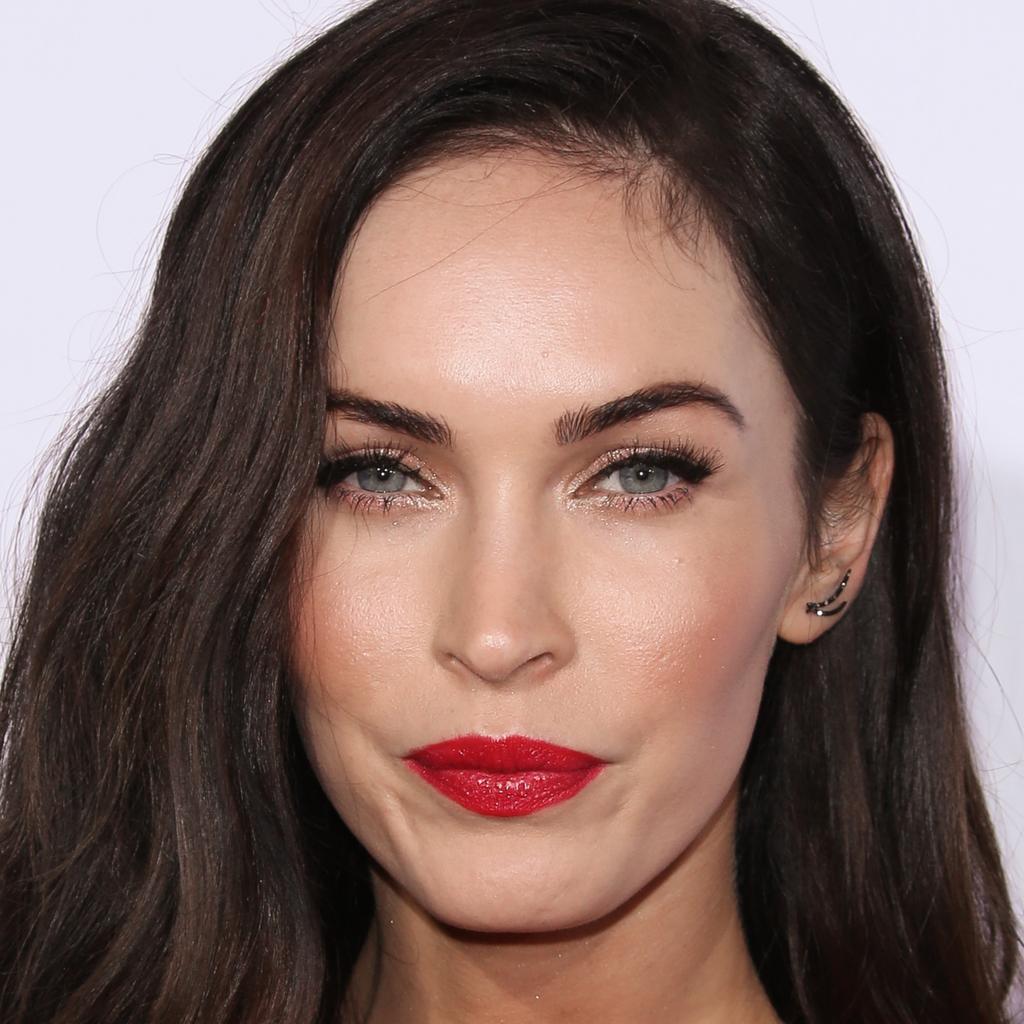}%
\newcommand{\varE}{var2268}\includegraphics[height=\h]{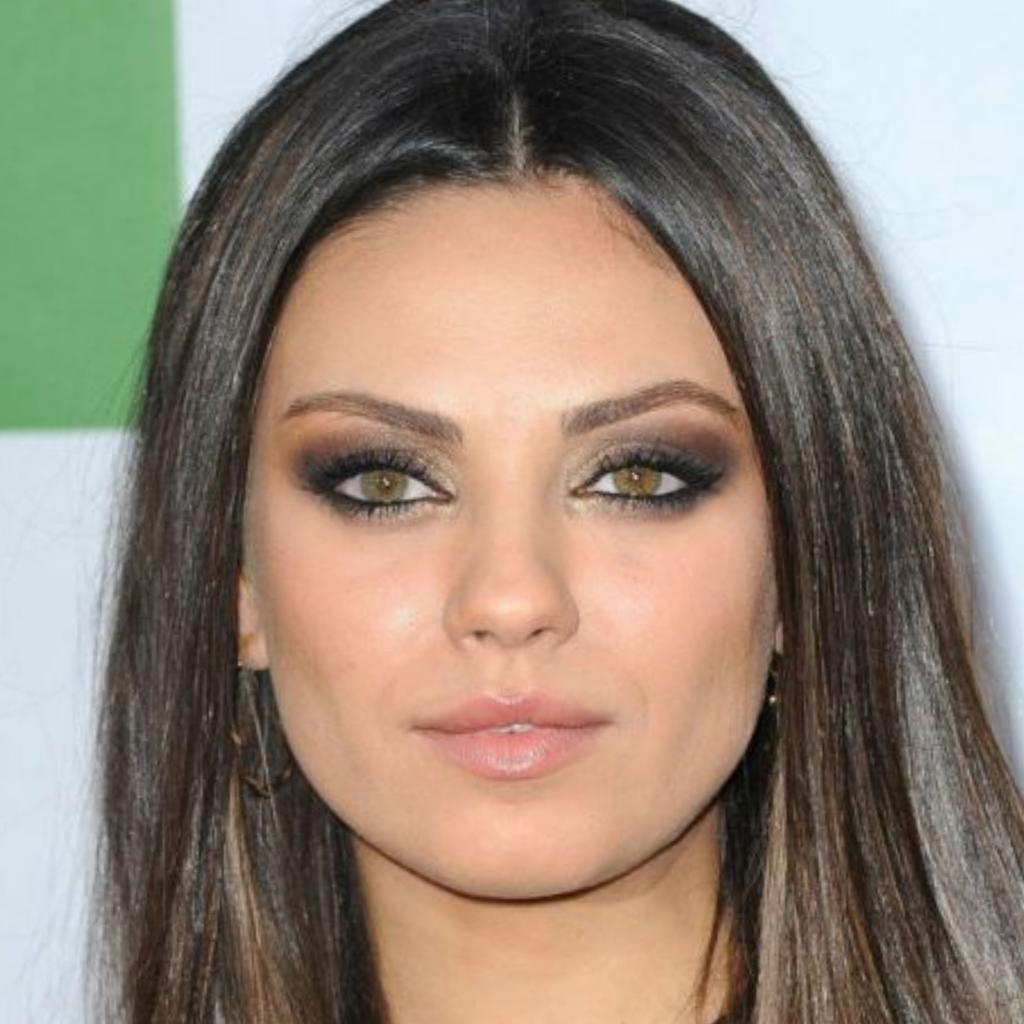}\vspace*{-3.5mm}\\
\begin{tikzpicture}\draw (0,0) -- (\linewidth,0);\end{tikzpicture}\vspace*{-0.5mm}\\%
\makebox[\hh]{\rotatebox[origin=l]{90}{\makebox[\h][c]{\hspace{-0.332\linewidth}\normalsize{Coarse styles from Source set}}}}\hspace{0.5mm}%
\newcommand{\seed}{seed888}
\includegraphics[height=\h]{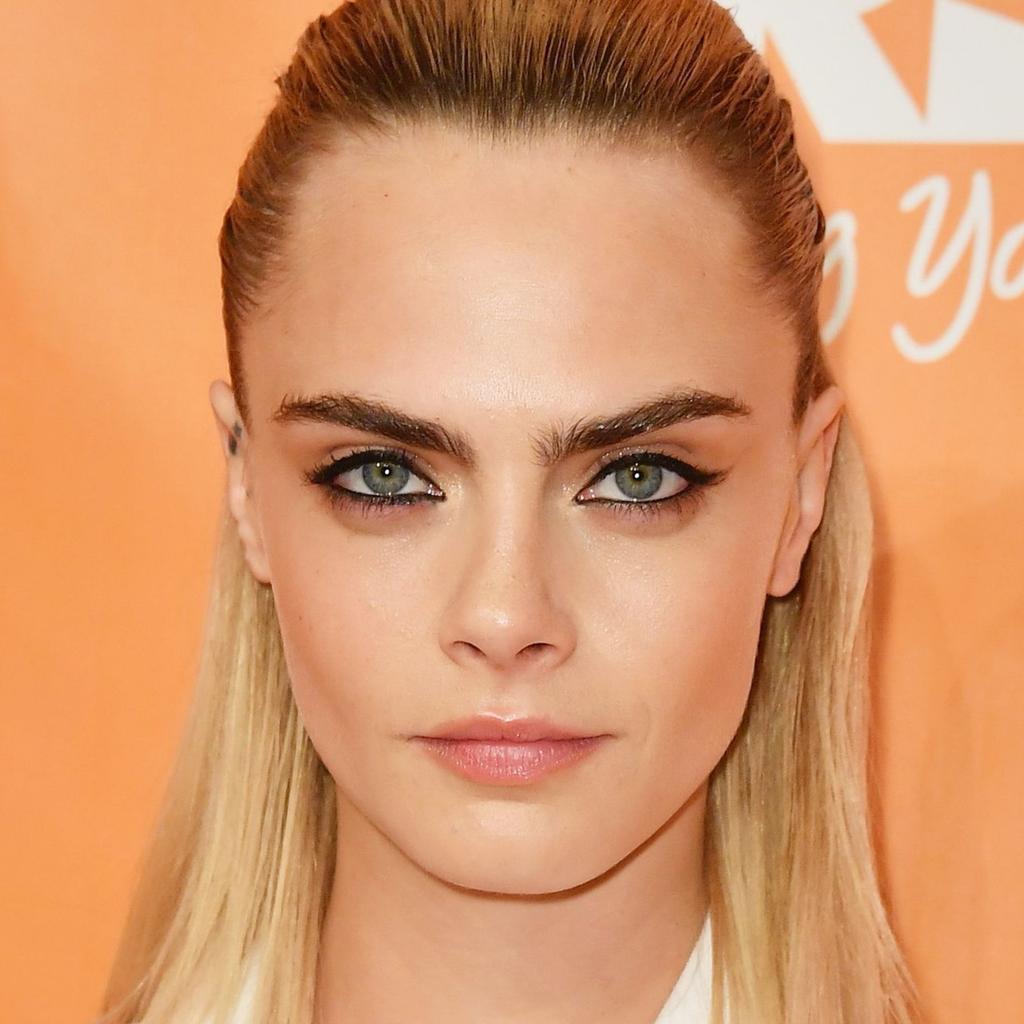}\hfill%
\includegraphics[height=\h]{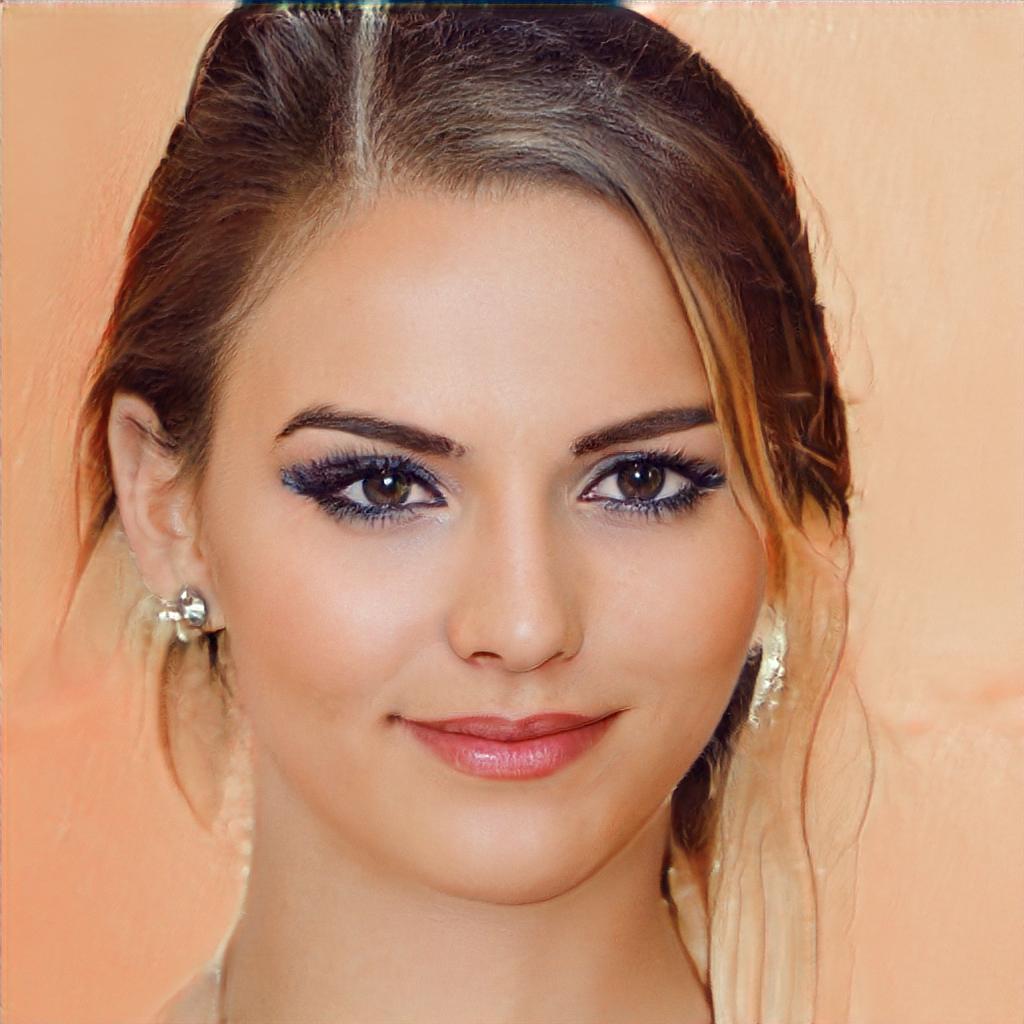}%
\includegraphics[height=\h]{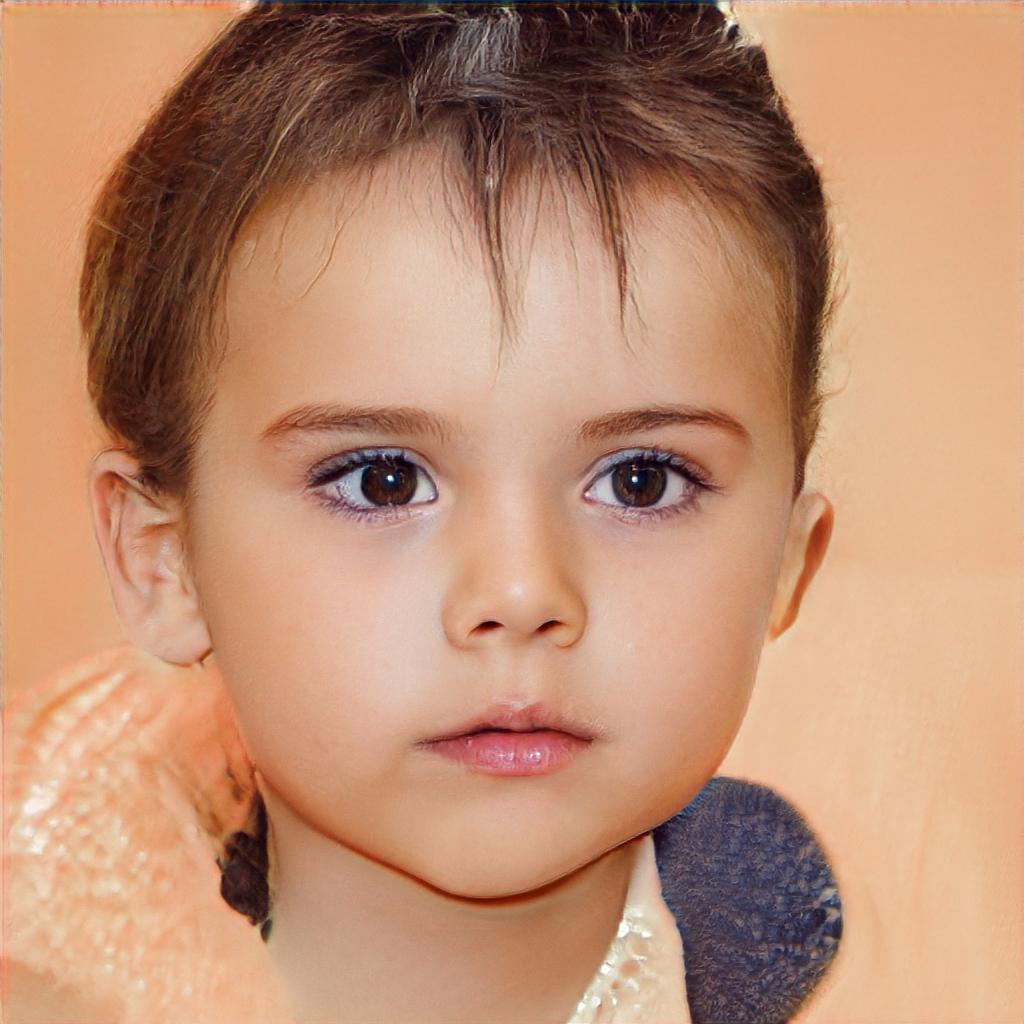}%
\includegraphics[height=\h]{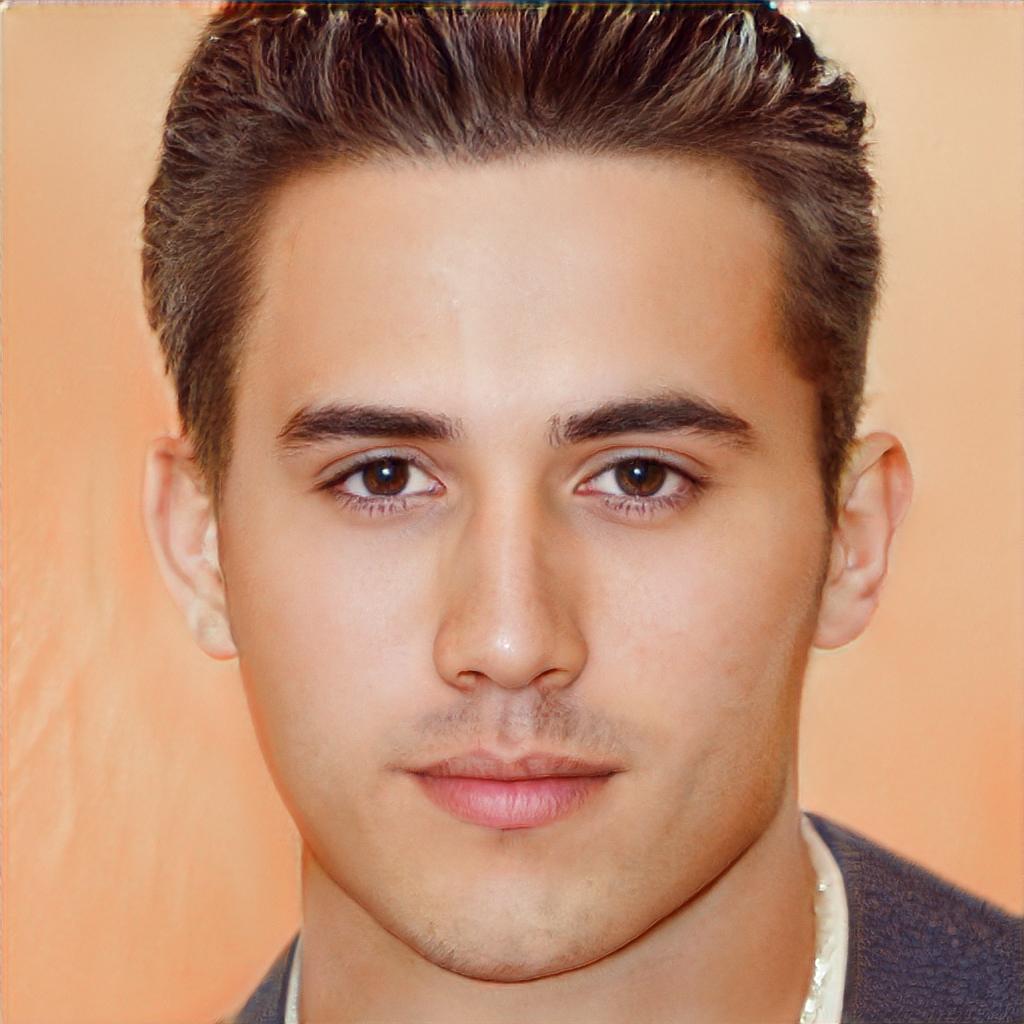}%
\includegraphics[height=\h]{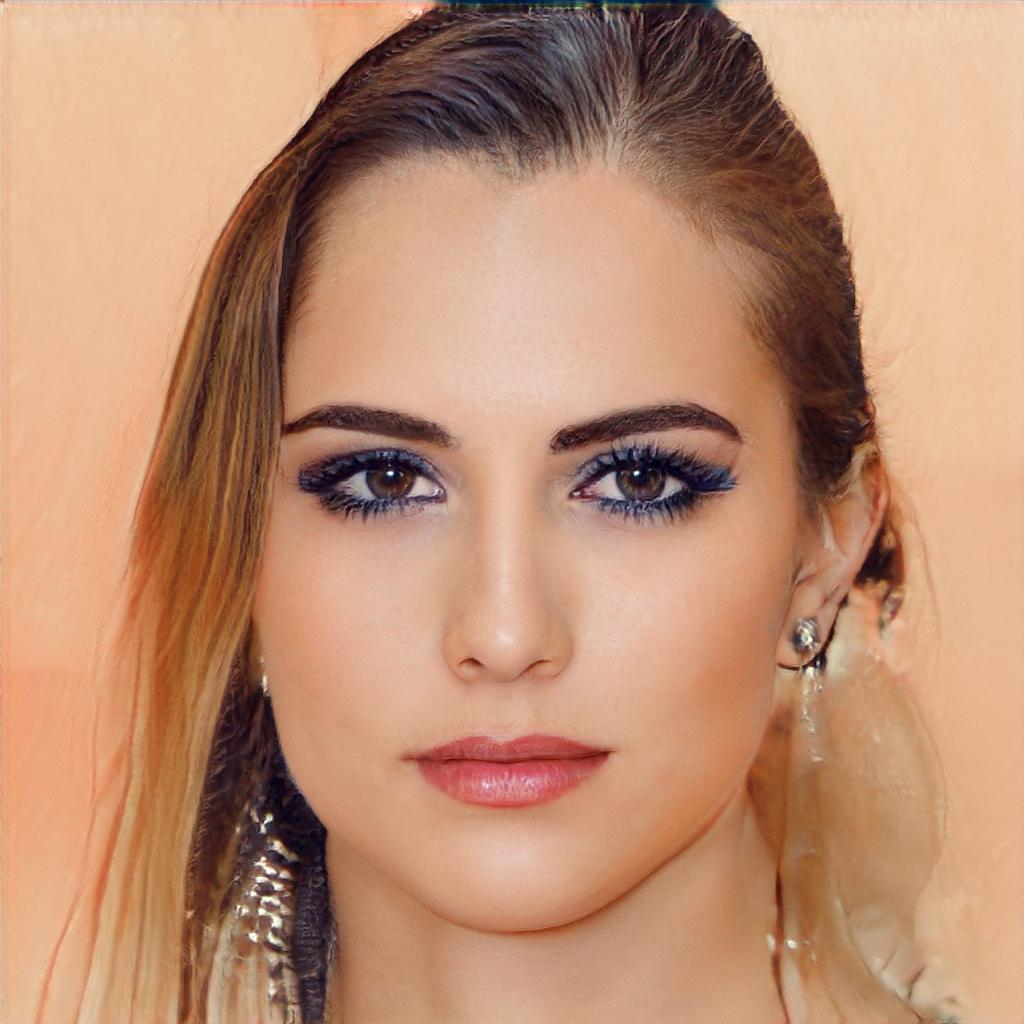}%
\includegraphics[height=\h]{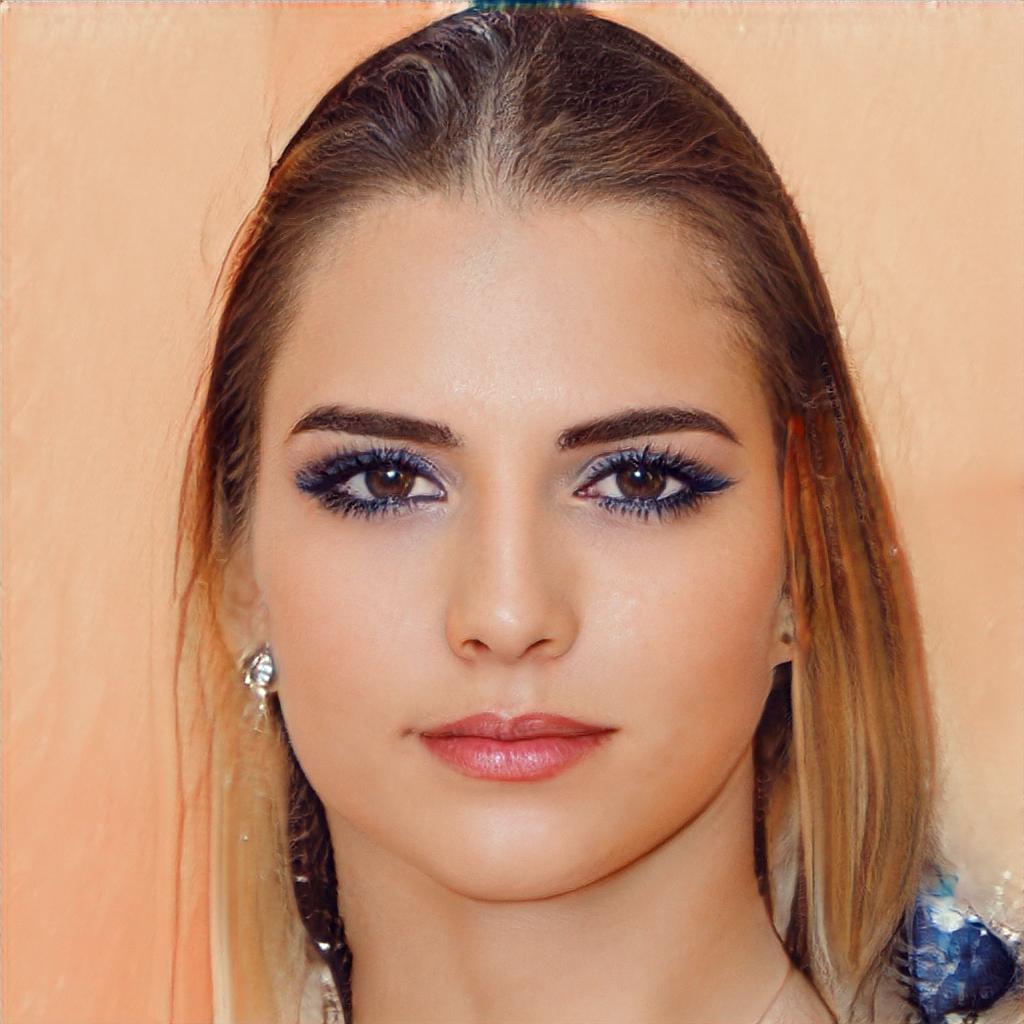}\vv\\
\makebox[\hh][c]{}\hspace{0.5mm}%
\renewcommand{\seed}{seed829}
\includegraphics[height=\h]{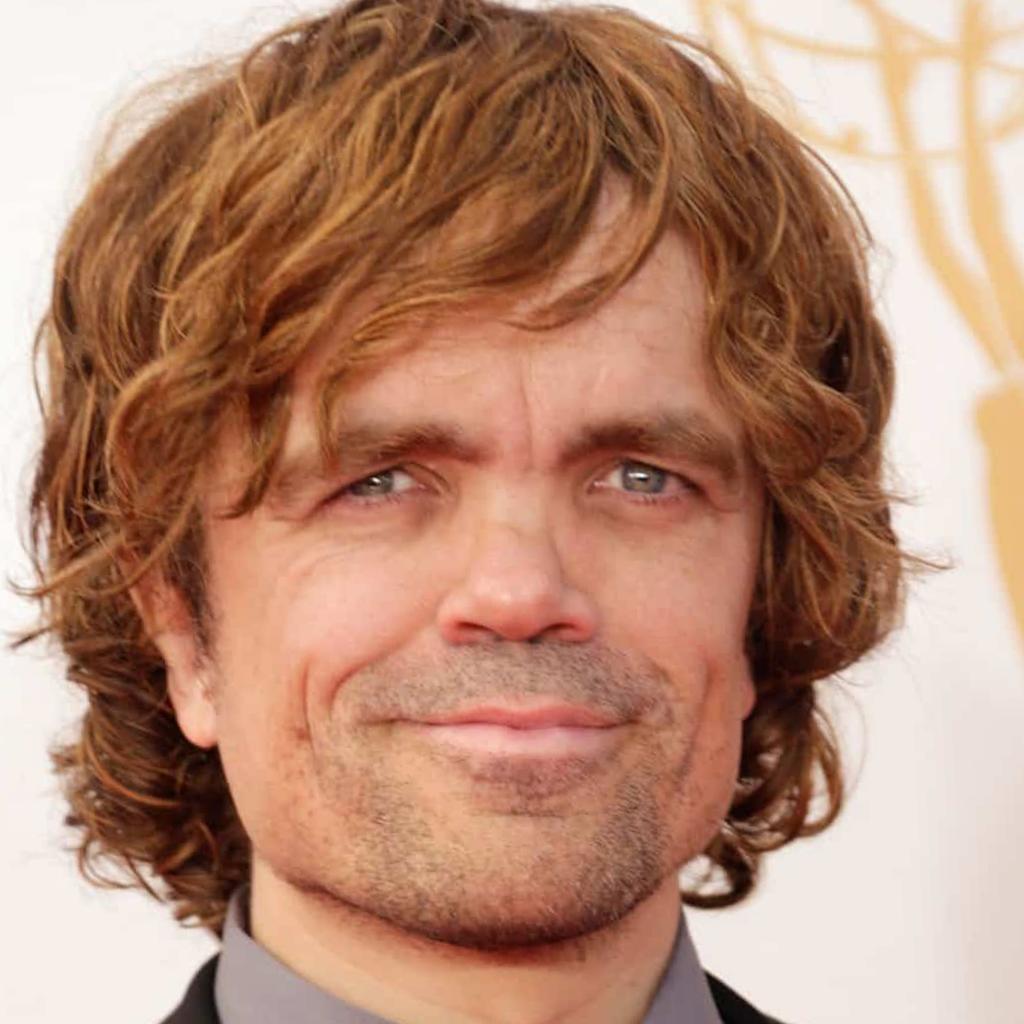}\hfill%
\includegraphics[height=\h]{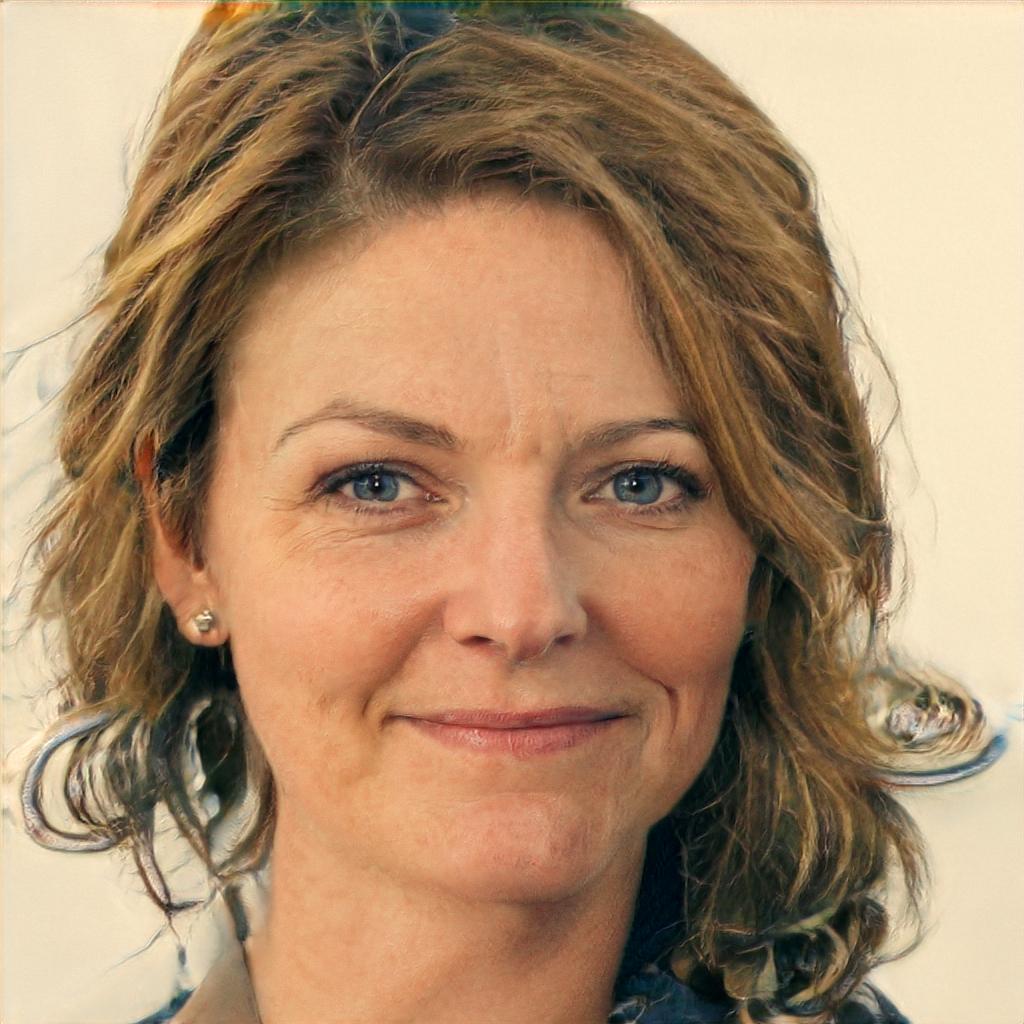}%
\includegraphics[height=\h]{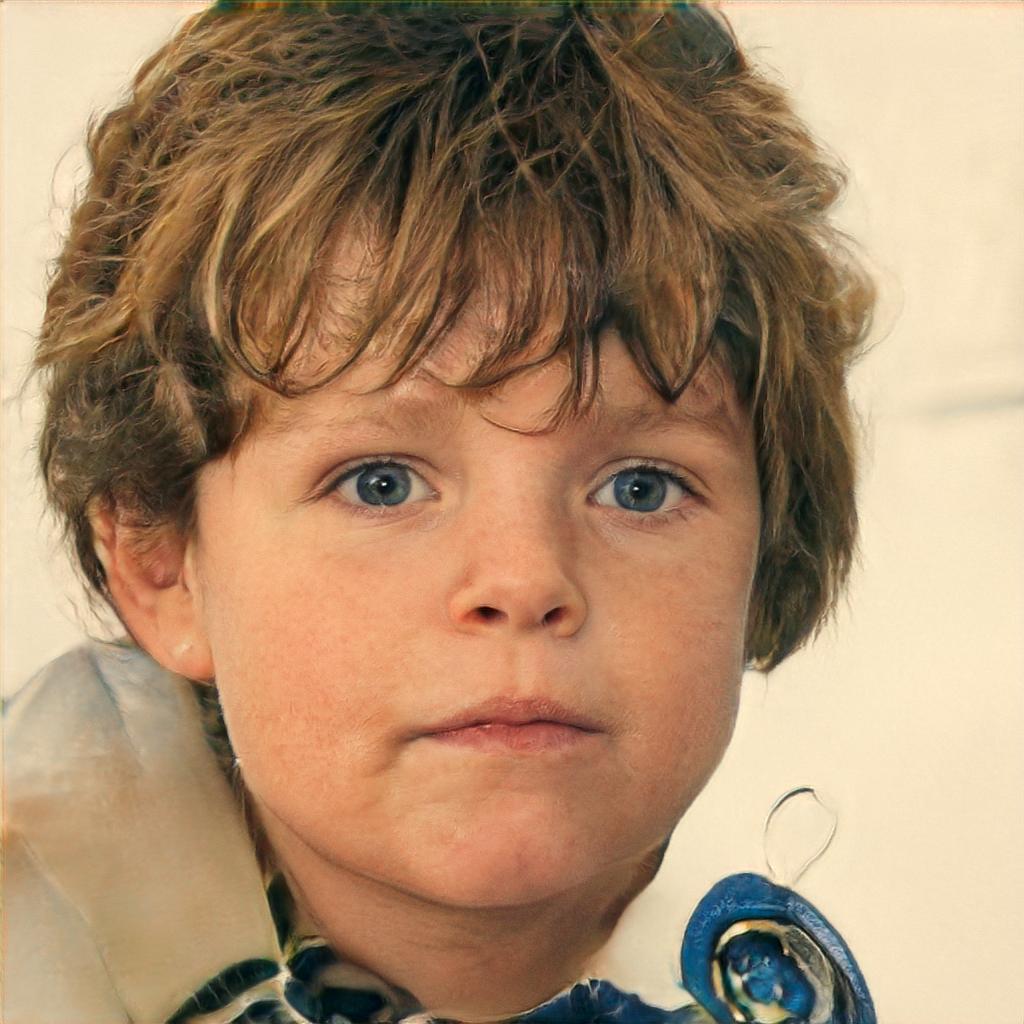}%
\includegraphics[height=\h]{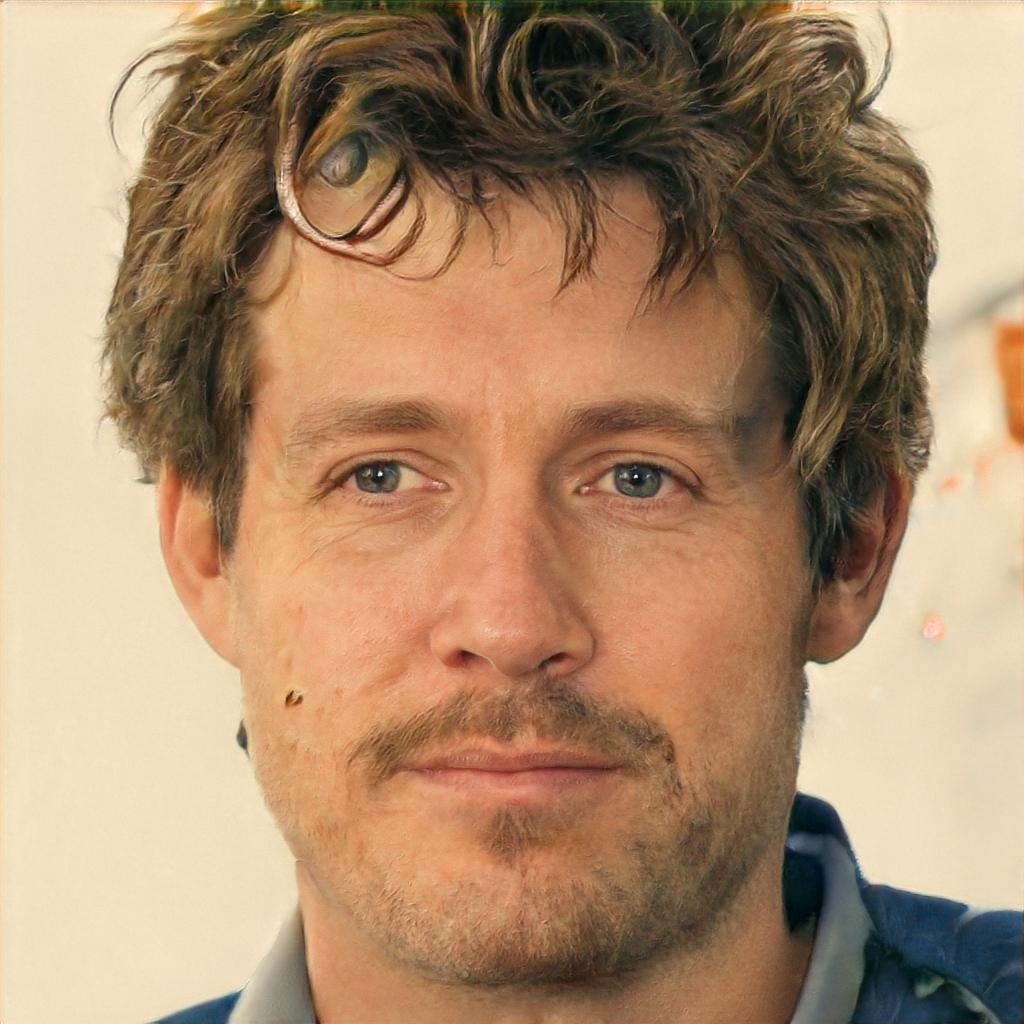}%
\includegraphics[height=\h]{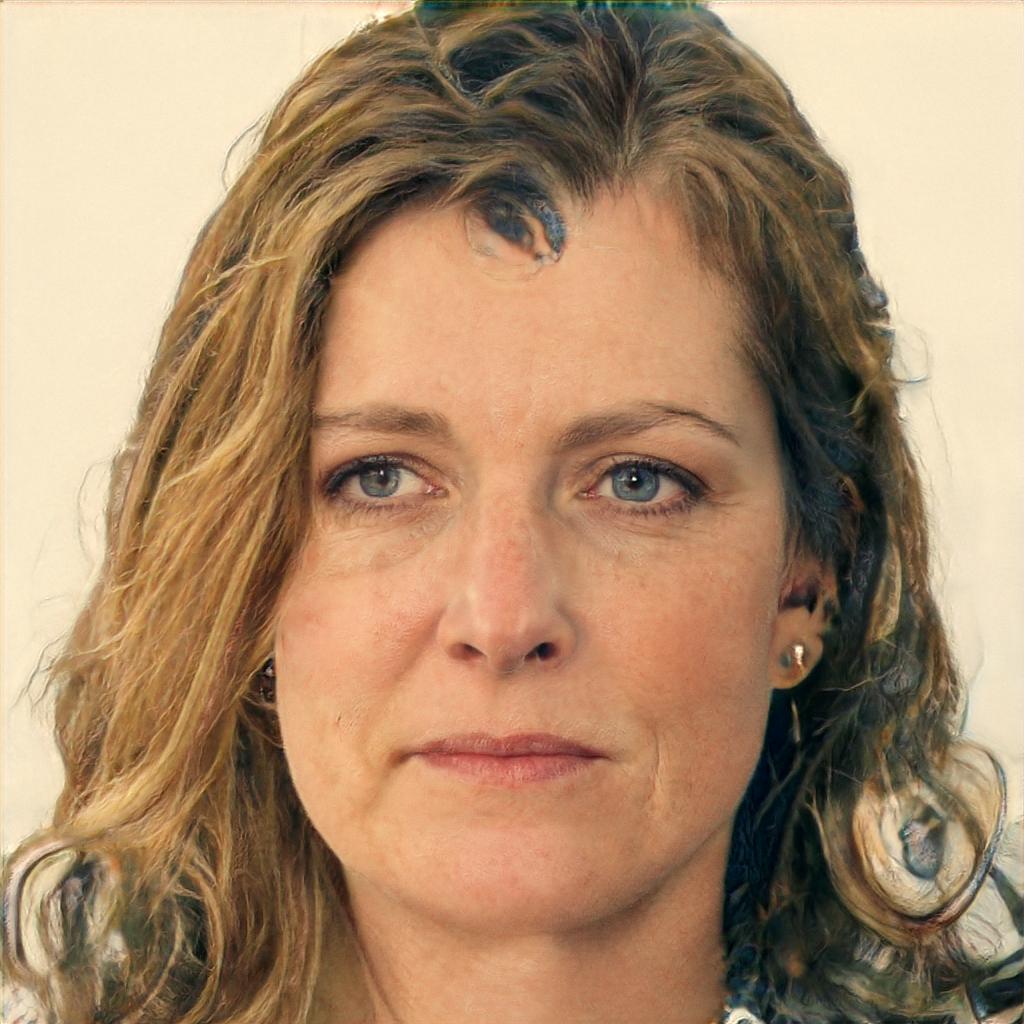}%
\includegraphics[height=\h]{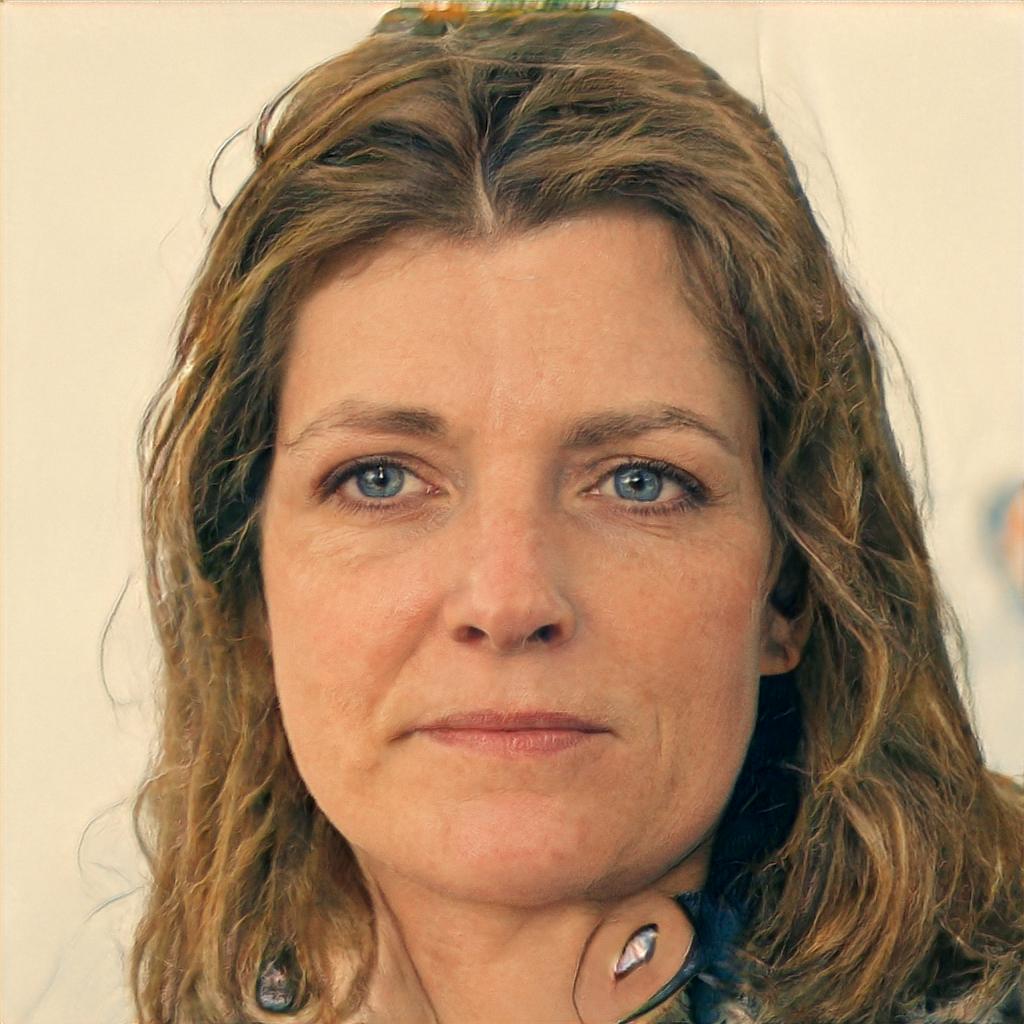}\vv\\
\makebox[\hh][c]{}\hspace{0.5mm}%
\renewcommand{\seed}{seed1898}
\includegraphics[height=\h]{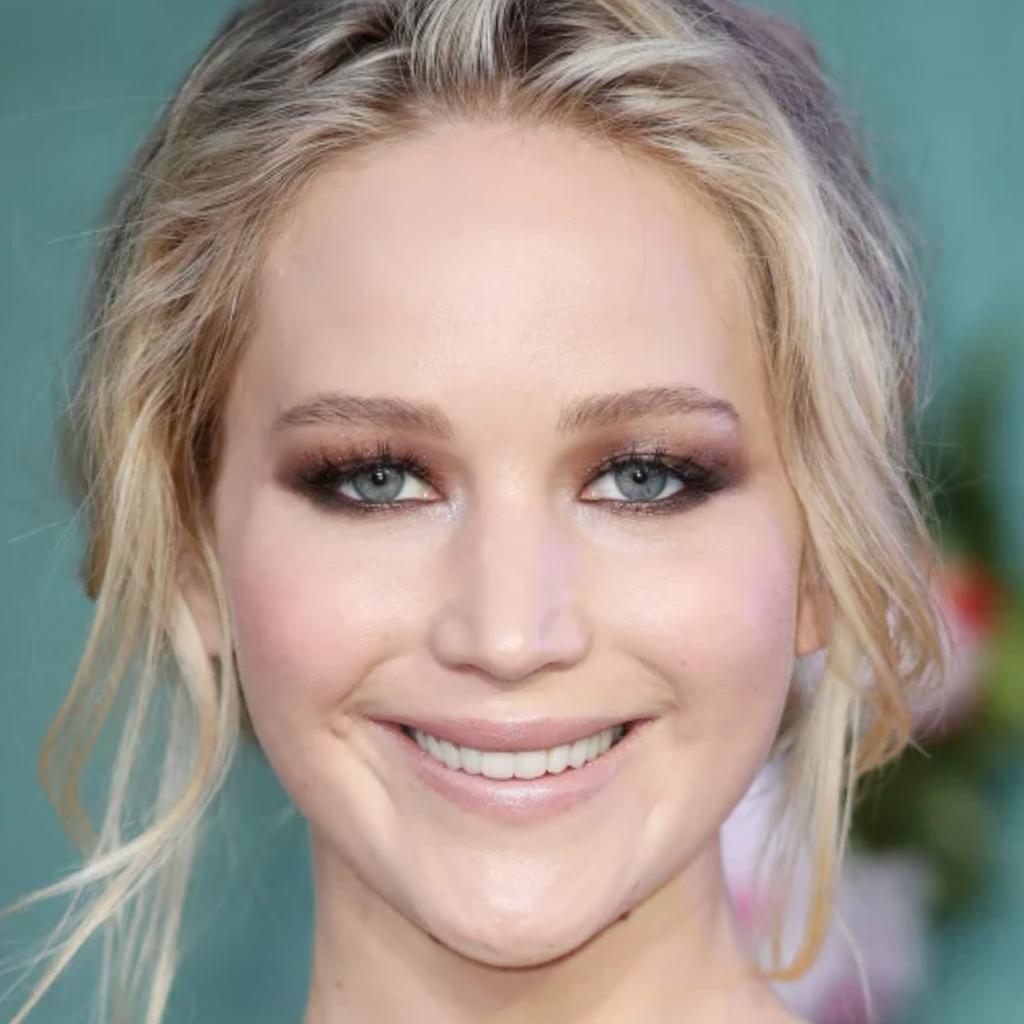}\hfill%
\includegraphics[height=\h]{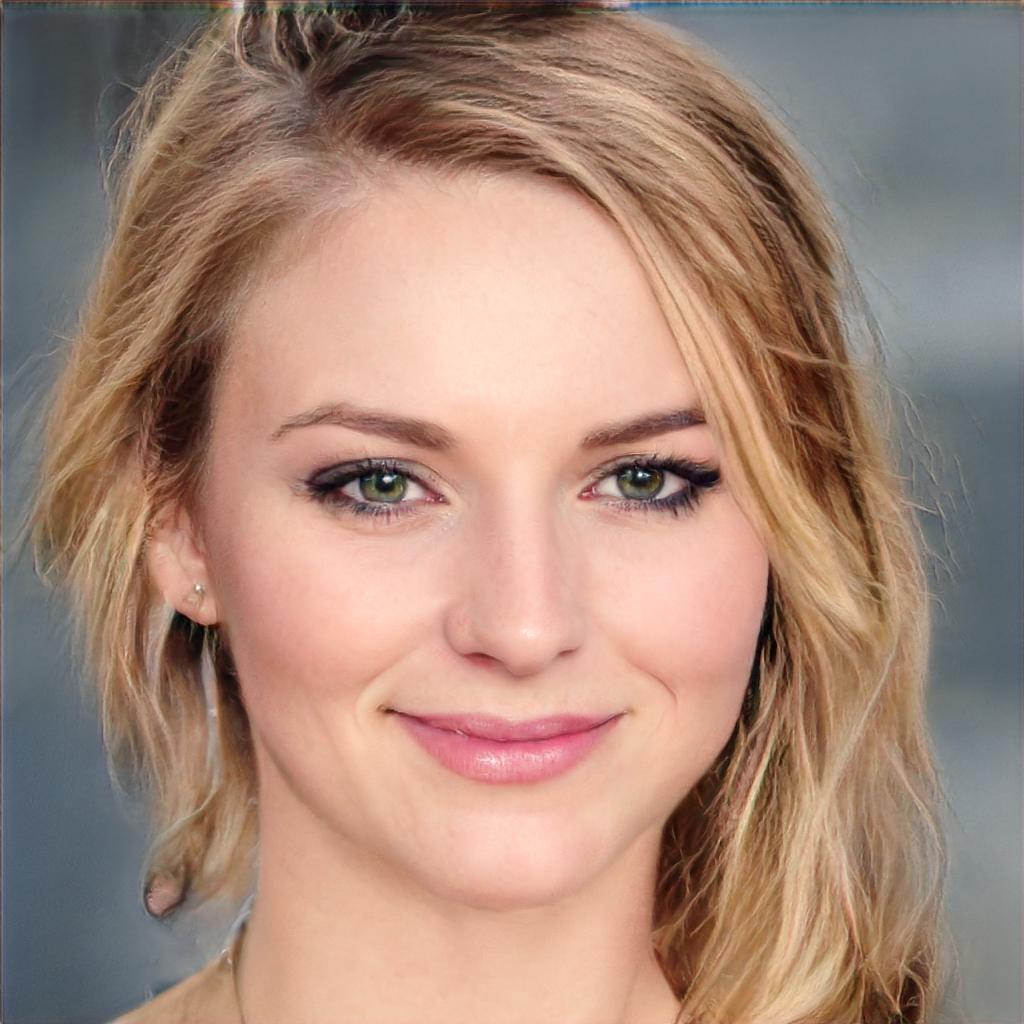}%
\includegraphics[height=\h]{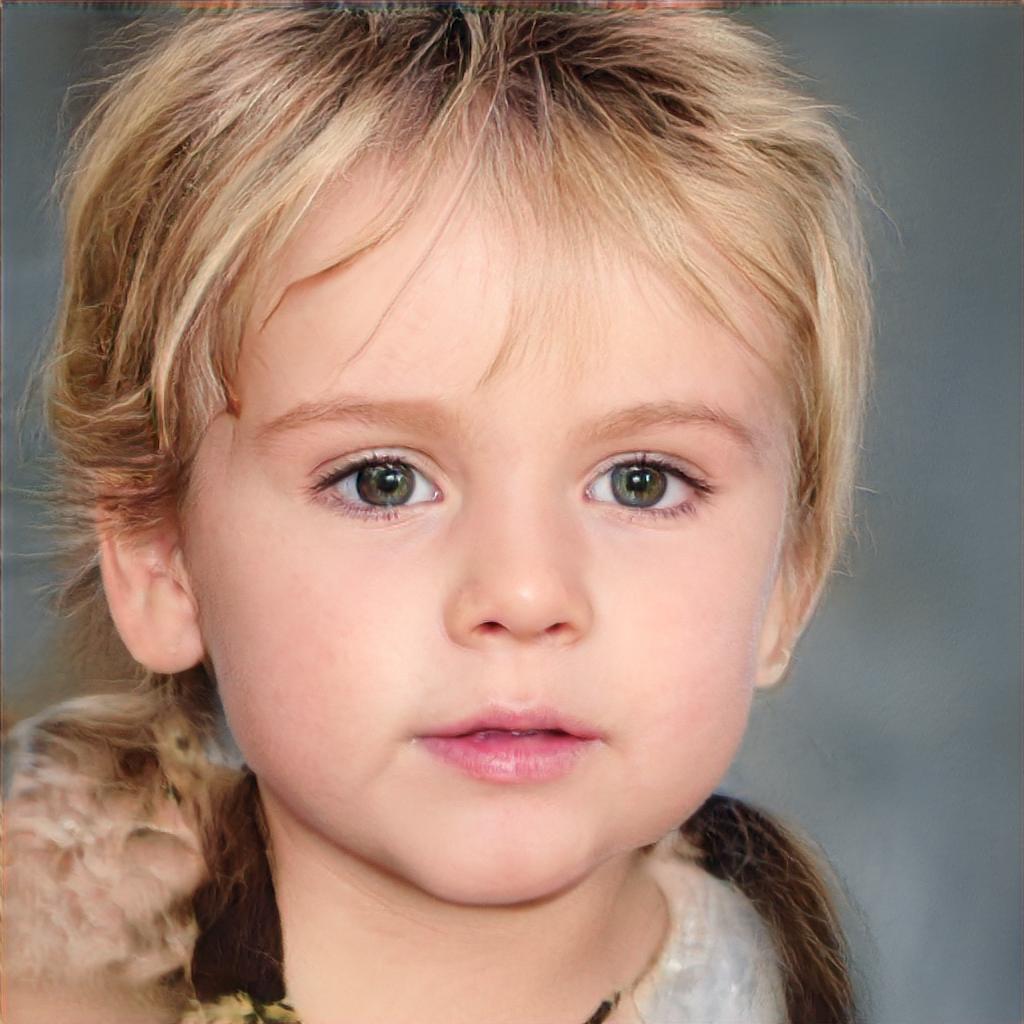}%
\includegraphics[height=\h]{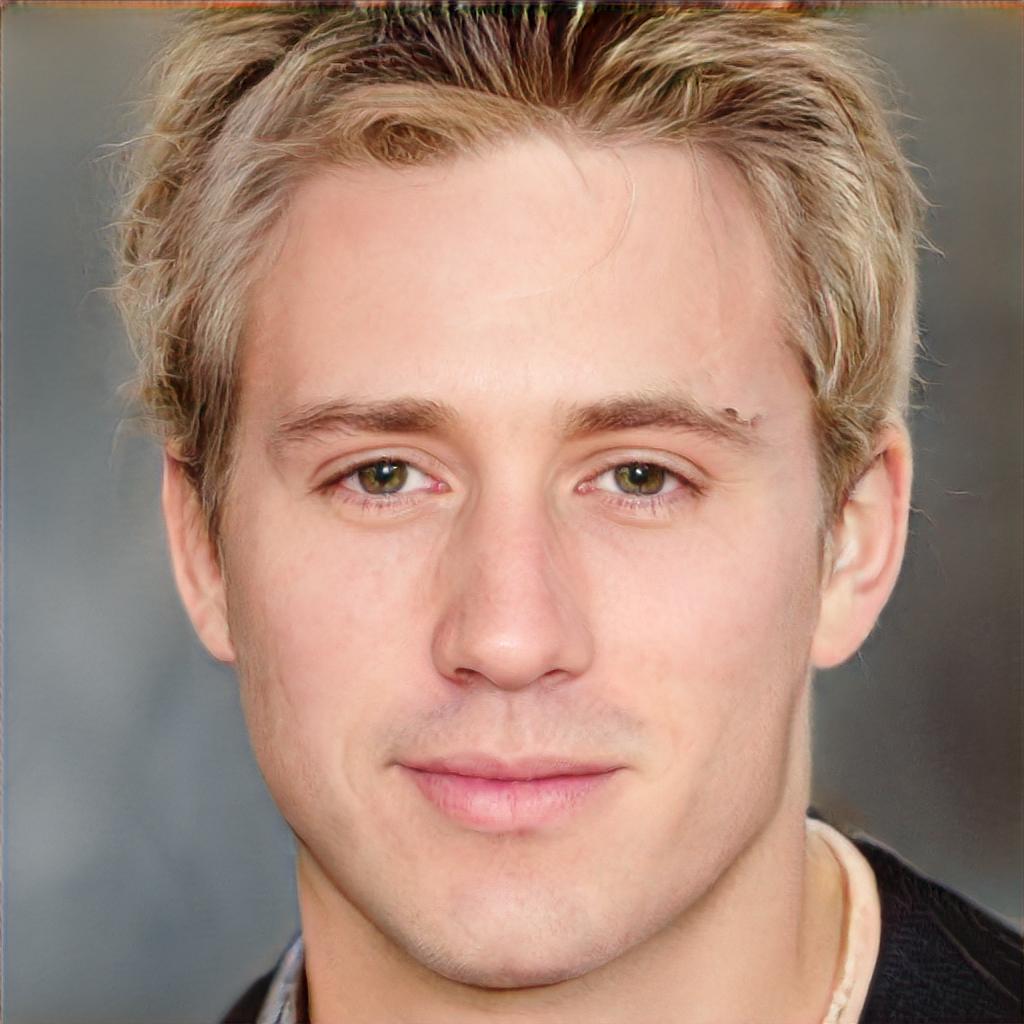}%
\includegraphics[height=\h]{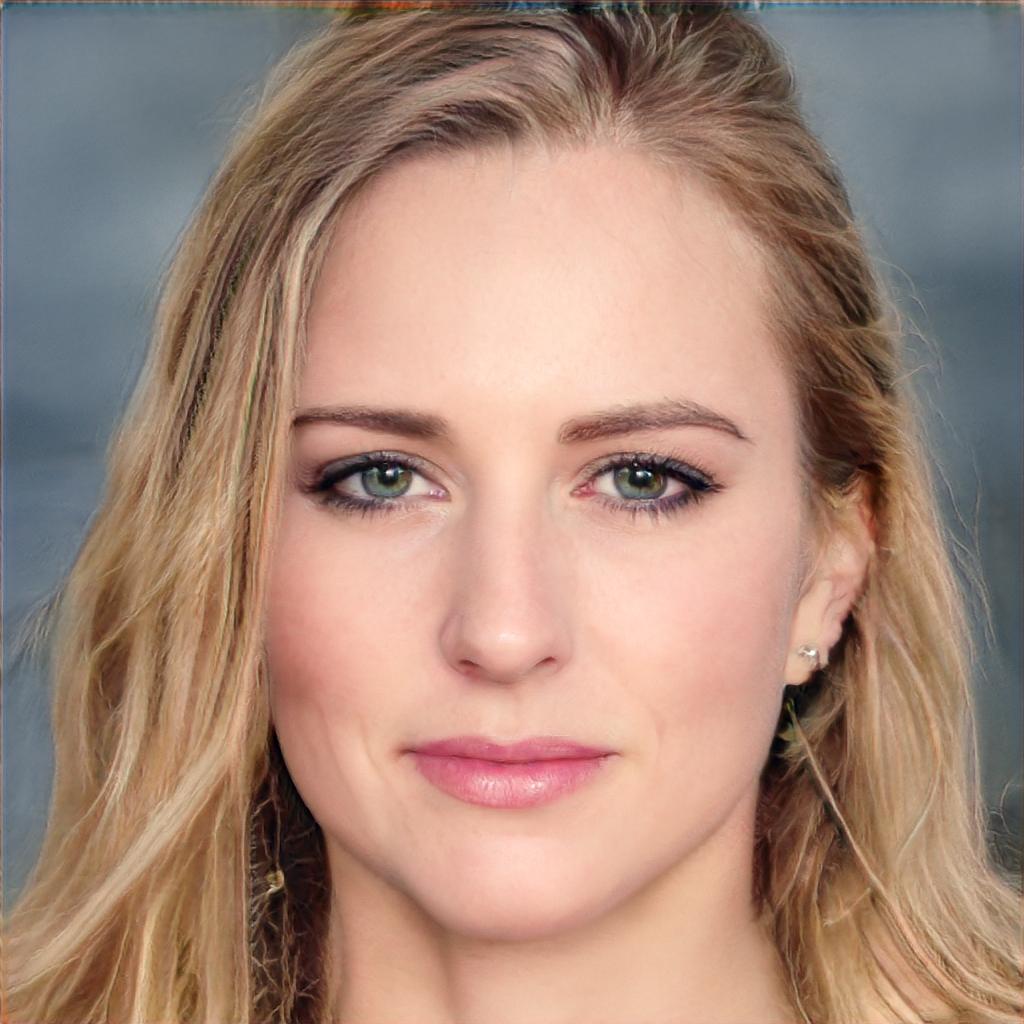}%
\includegraphics[height=\h]{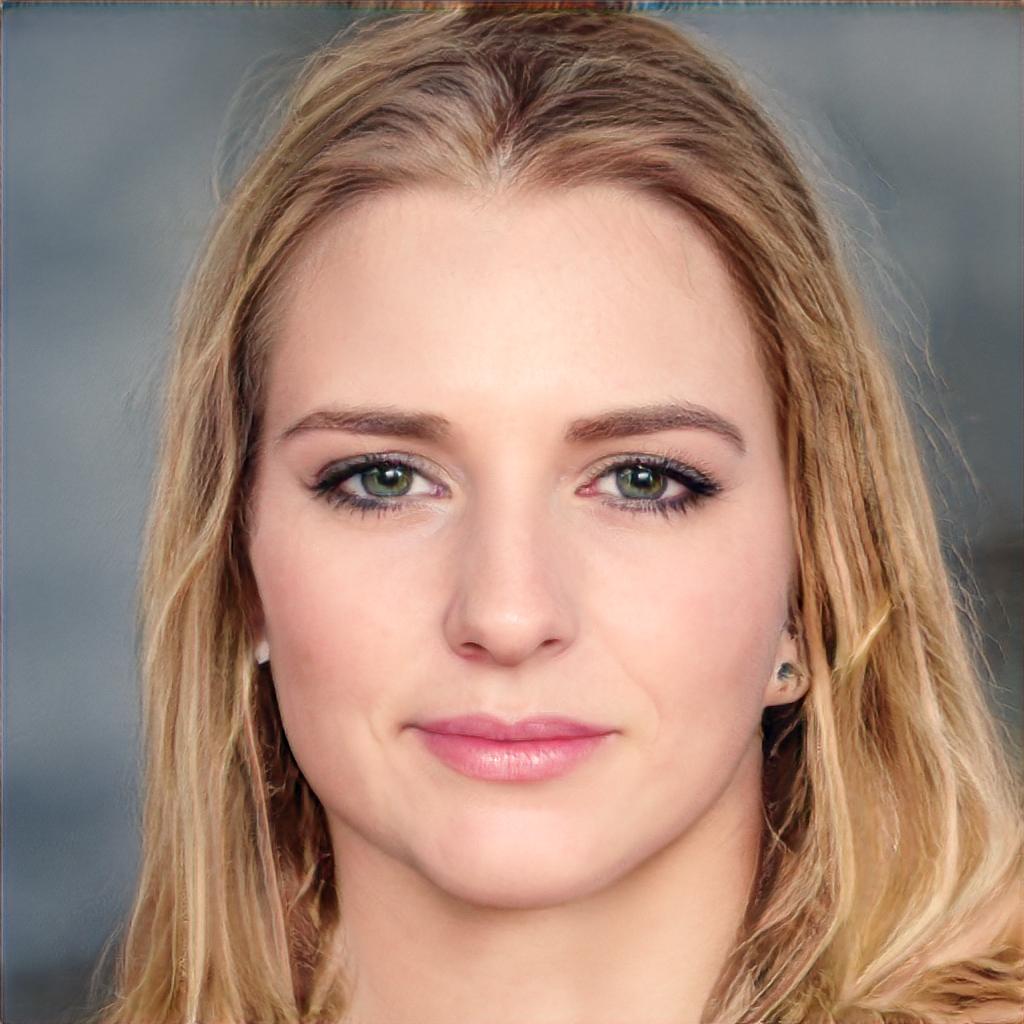}\vspace*{-3.5mm}\\
\begin{tikzpicture}\draw (0,0) -- (\linewidth,0);\end{tikzpicture}\vspace*{-0.5mm}\\%
\makebox[\hh]{\rotatebox[origin=l]{90}{\makebox[\h][c]{\hspace{-\h}\normalsize{Middle styles from Source set}}}}\hspace{0.5mm}%
\renewcommand{\seed}{seed1733}
\includegraphics[height=\h]{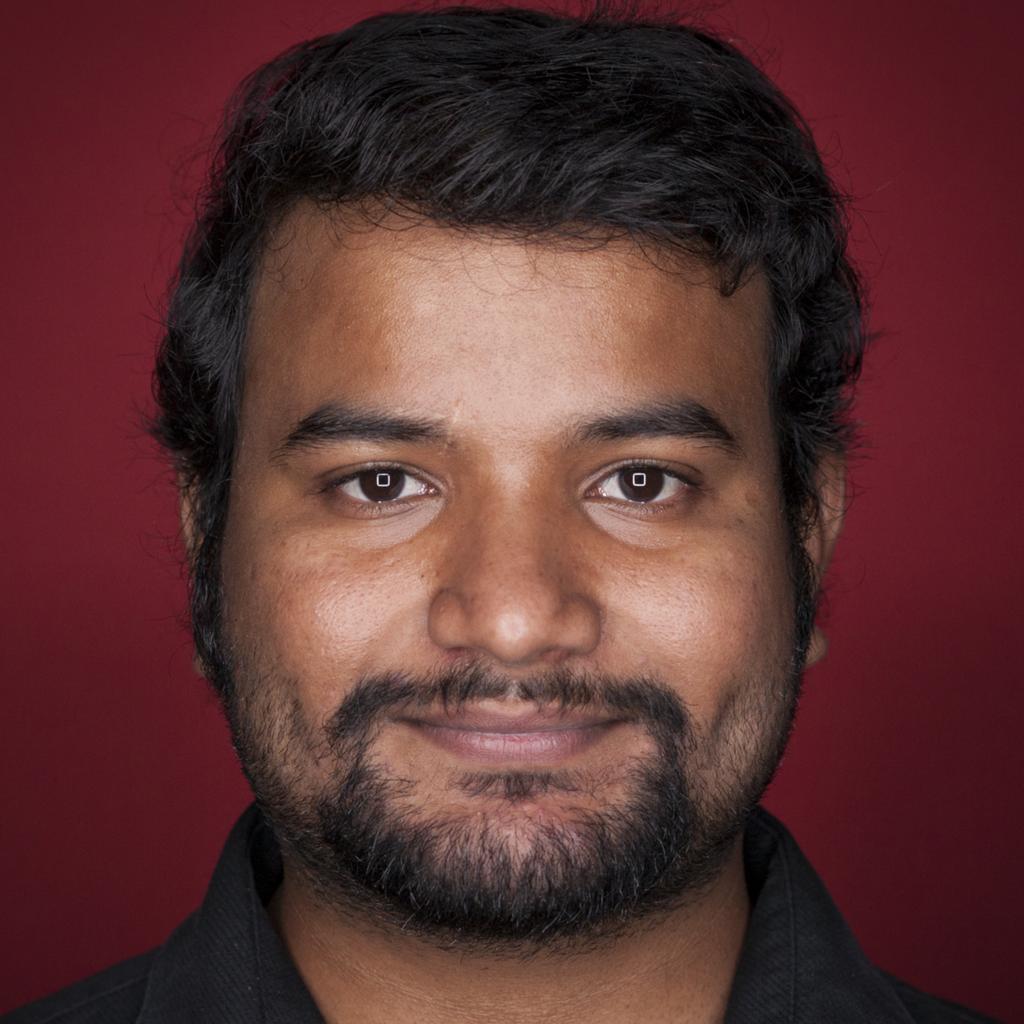}\hfill%
\includegraphics[height=\h]{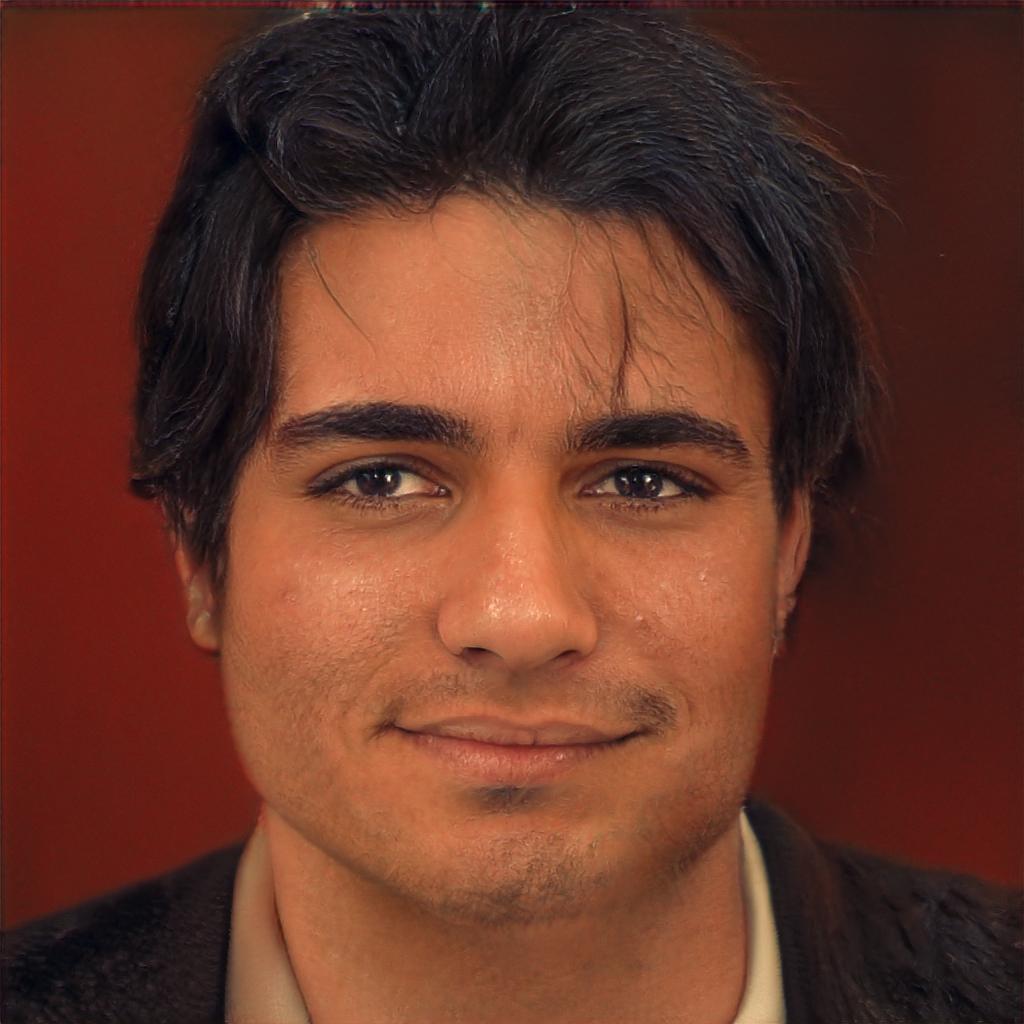}%
\includegraphics[height=\h]{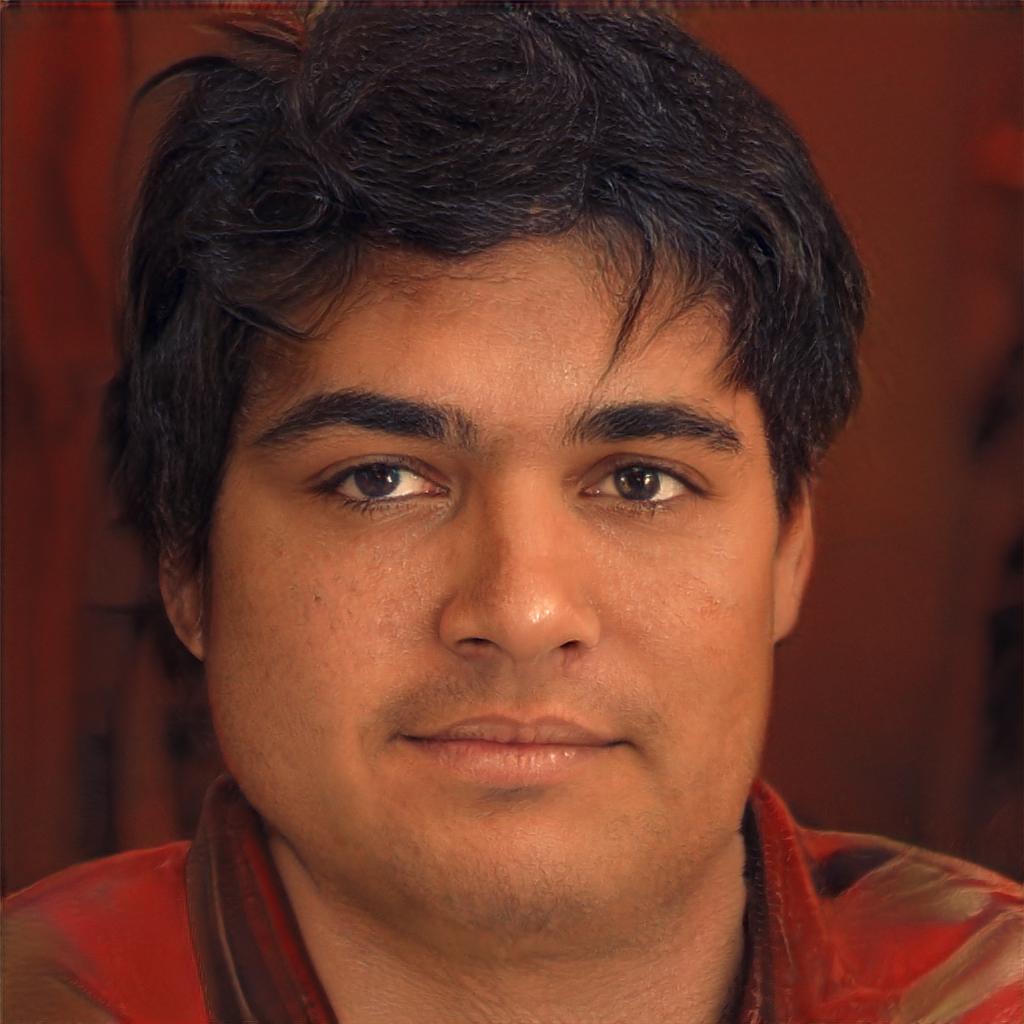}%
\includegraphics[height=\h]{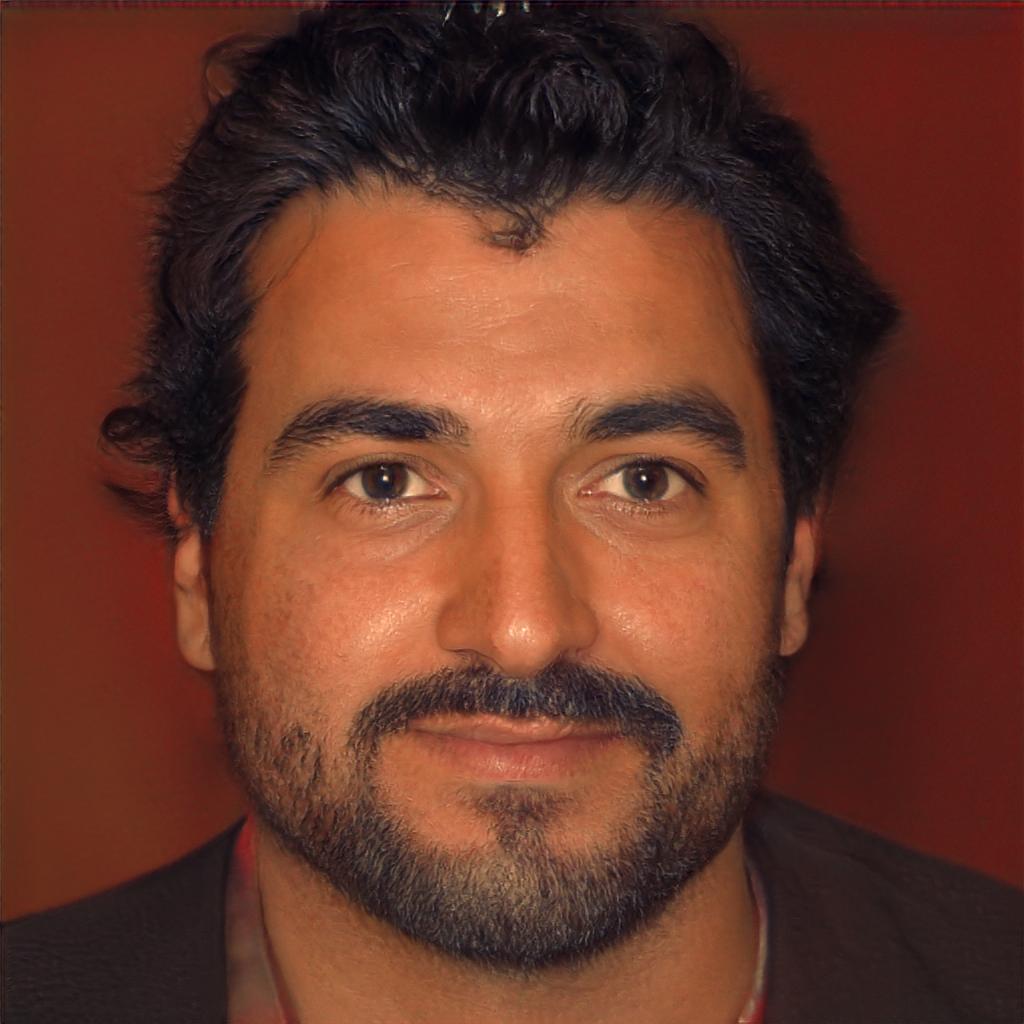}%
\includegraphics[height=\h]{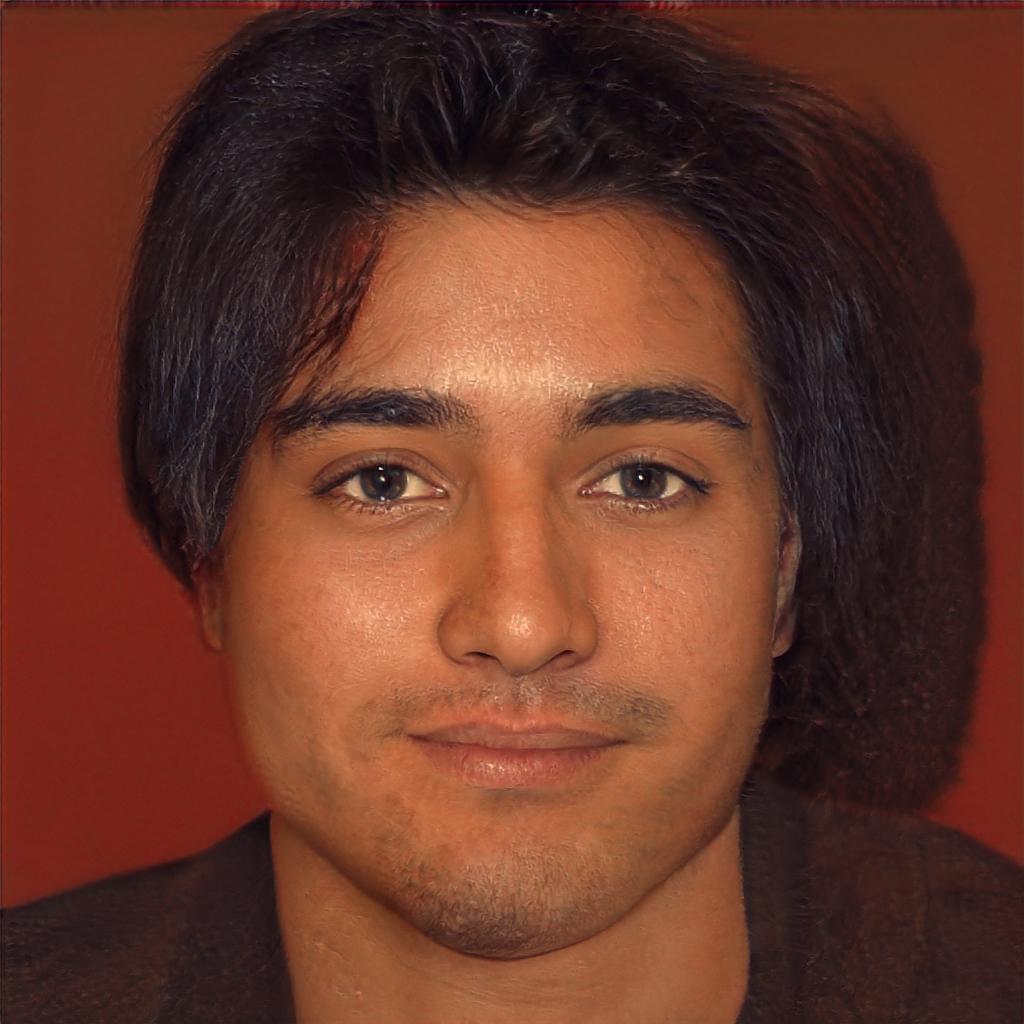}%
\includegraphics[height=\h]{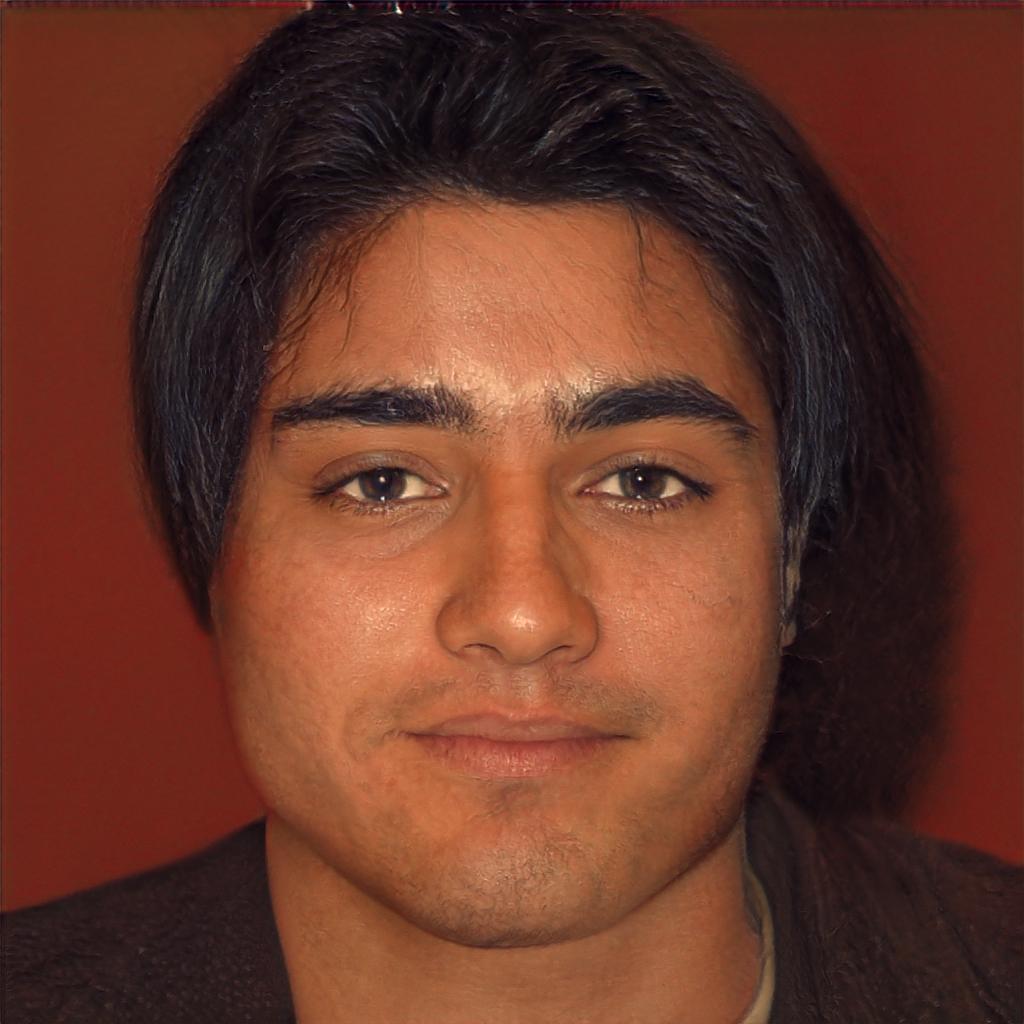}\vv\\
\makebox[\hh][c]{}\hspace{0.5mm}%
\renewcommand{\seed}{seed1614}
\includegraphics[height=\h]{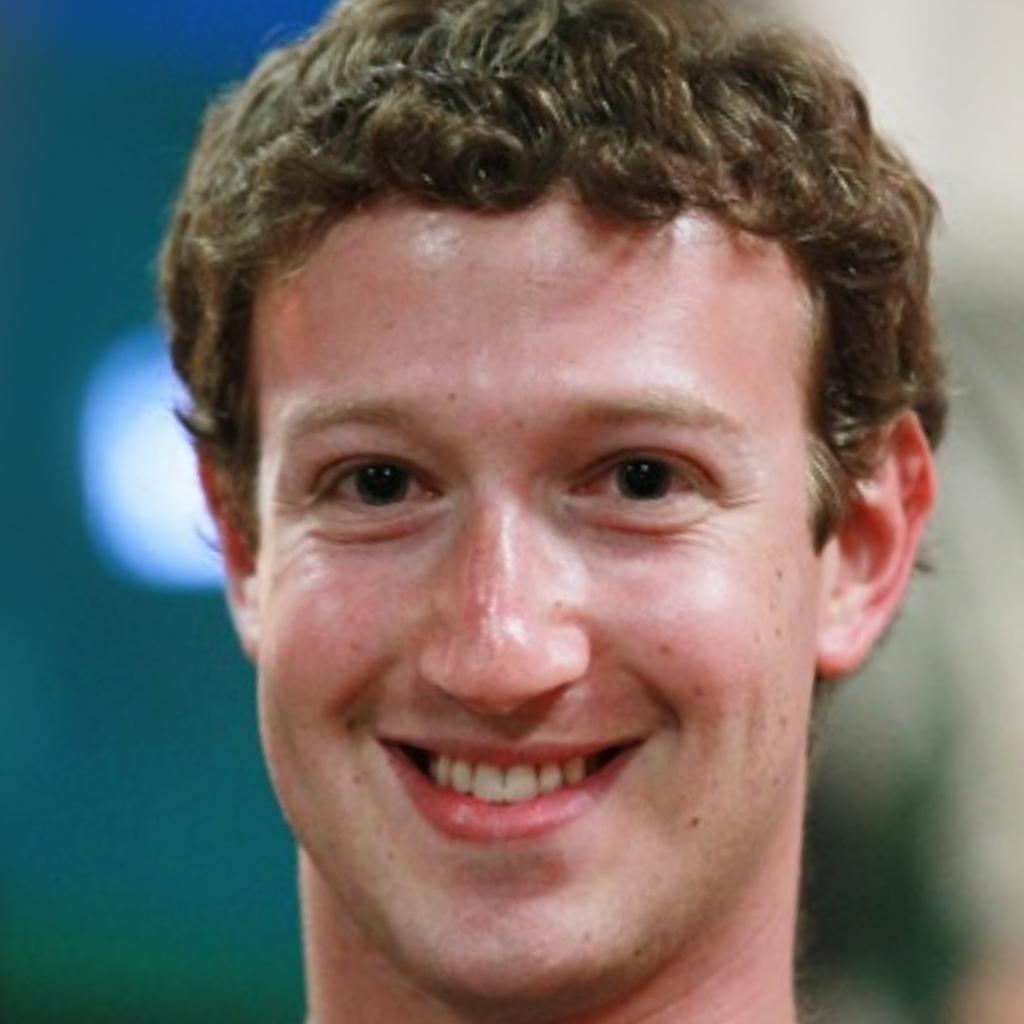}\hfill%
\includegraphics[height=\h]{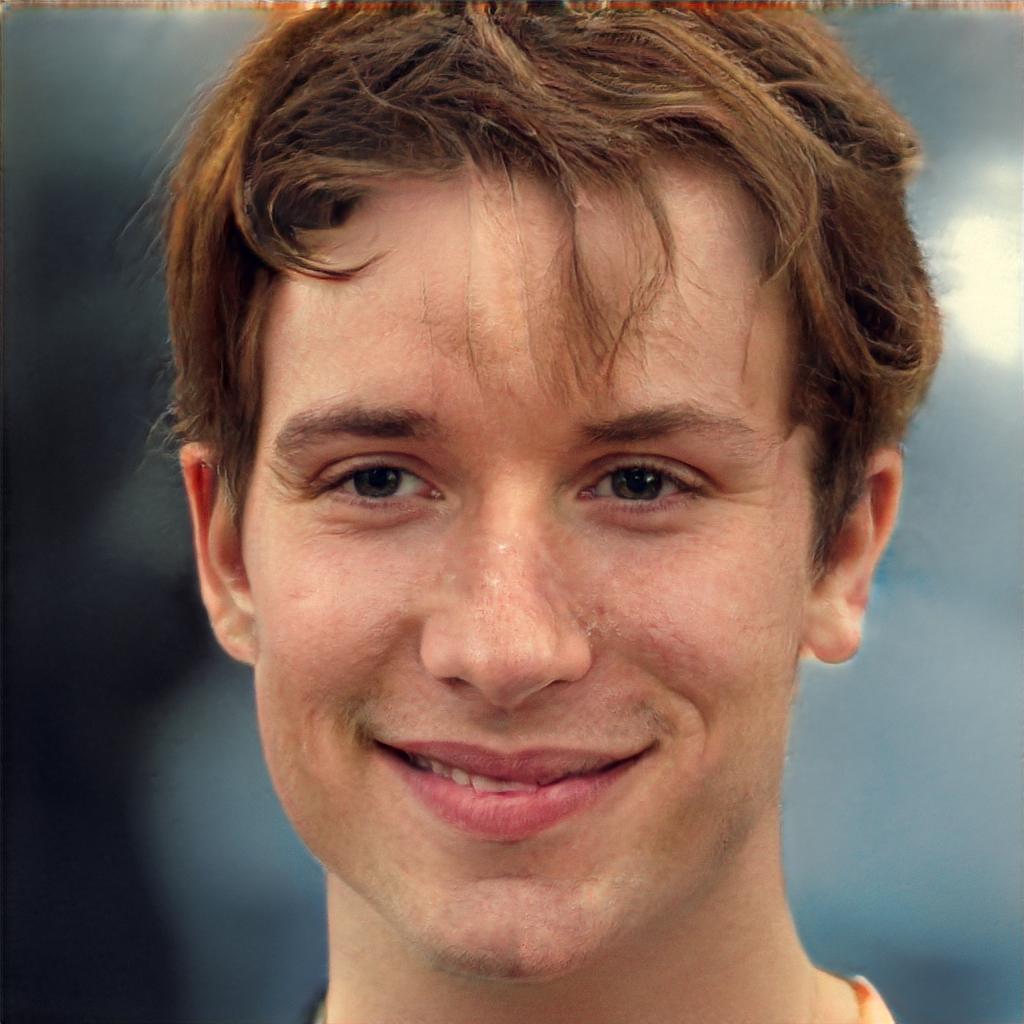}%
\includegraphics[height=\h]{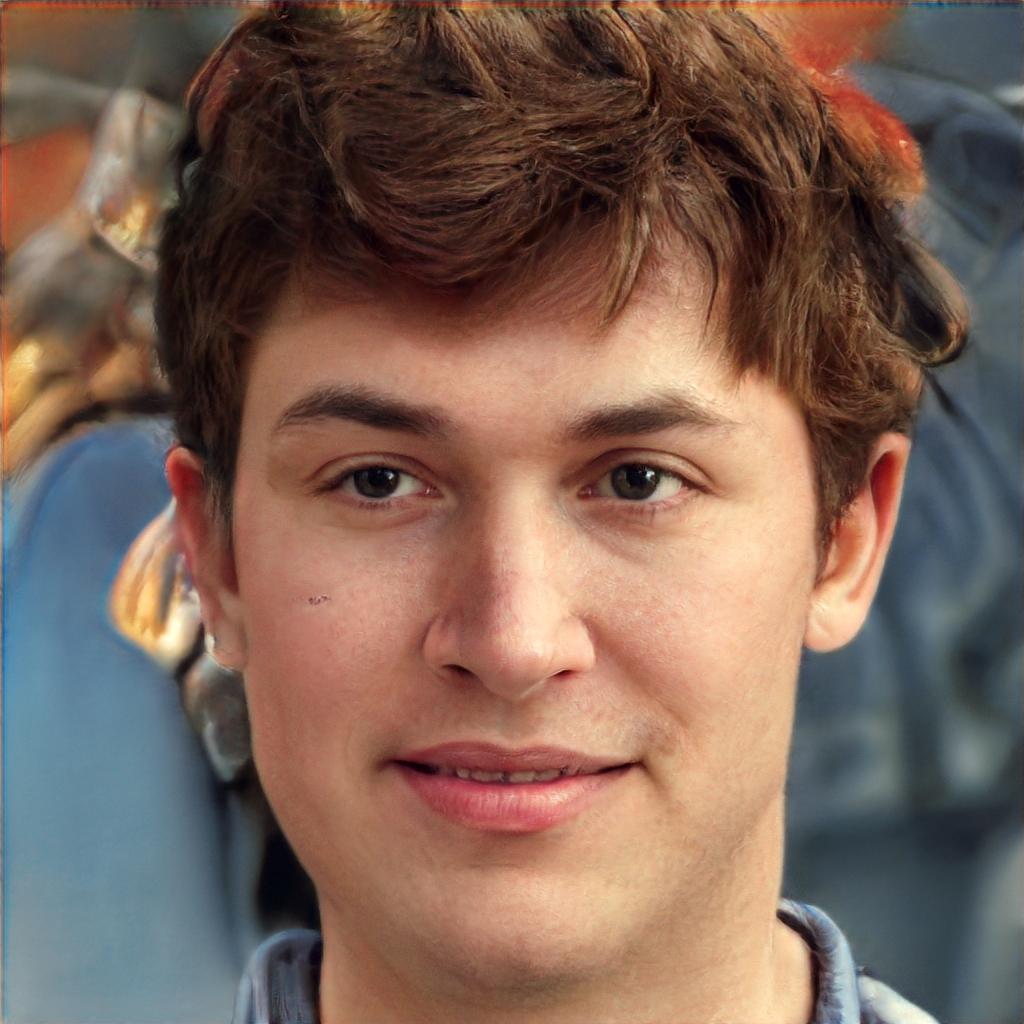}%
\includegraphics[height=\h]{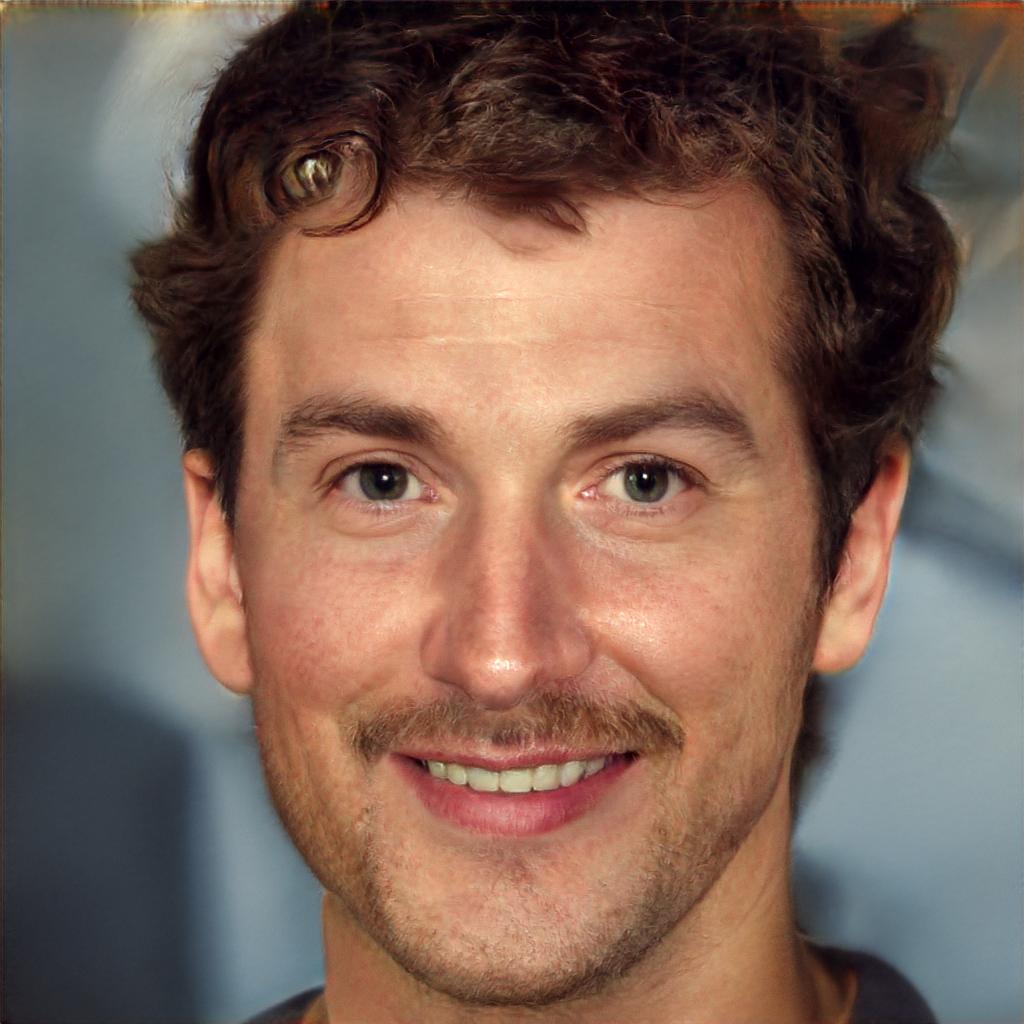}%
\includegraphics[height=\h]{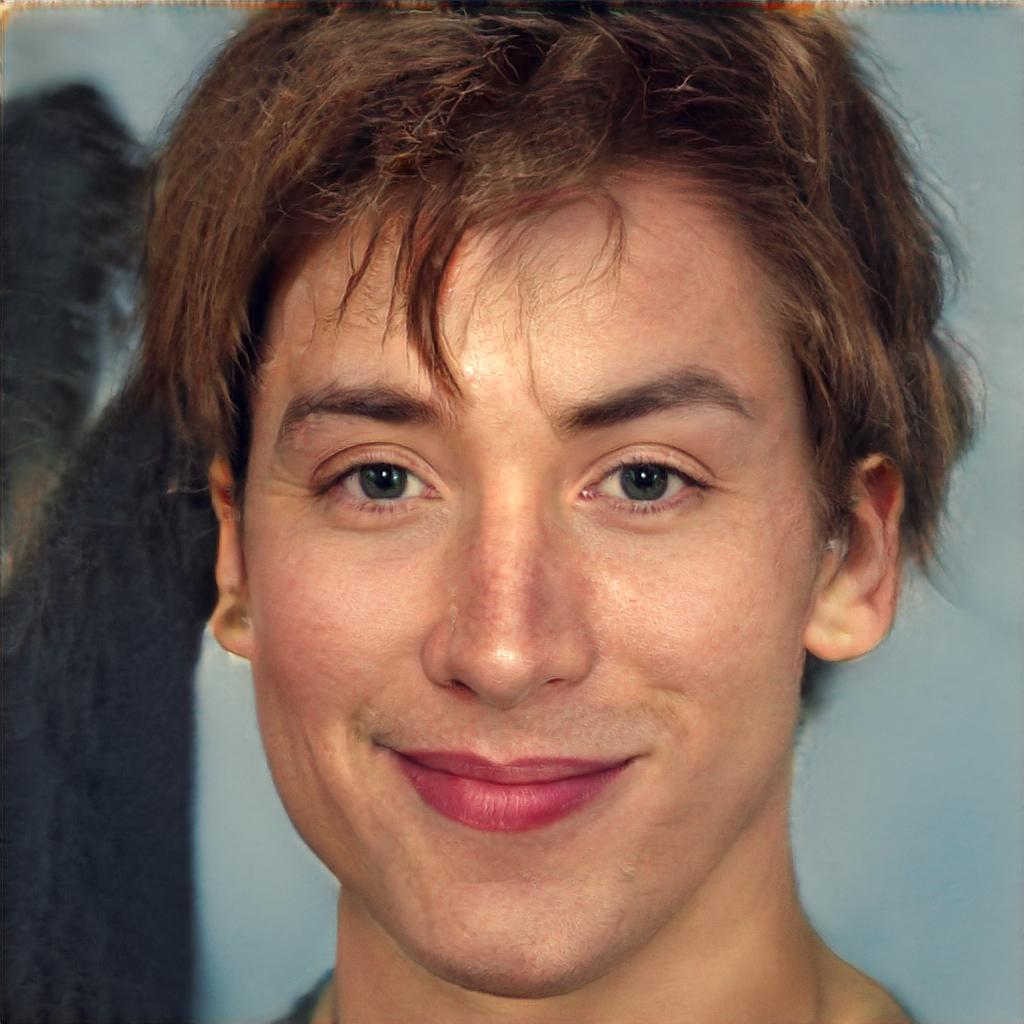}%
\includegraphics[height=\h]{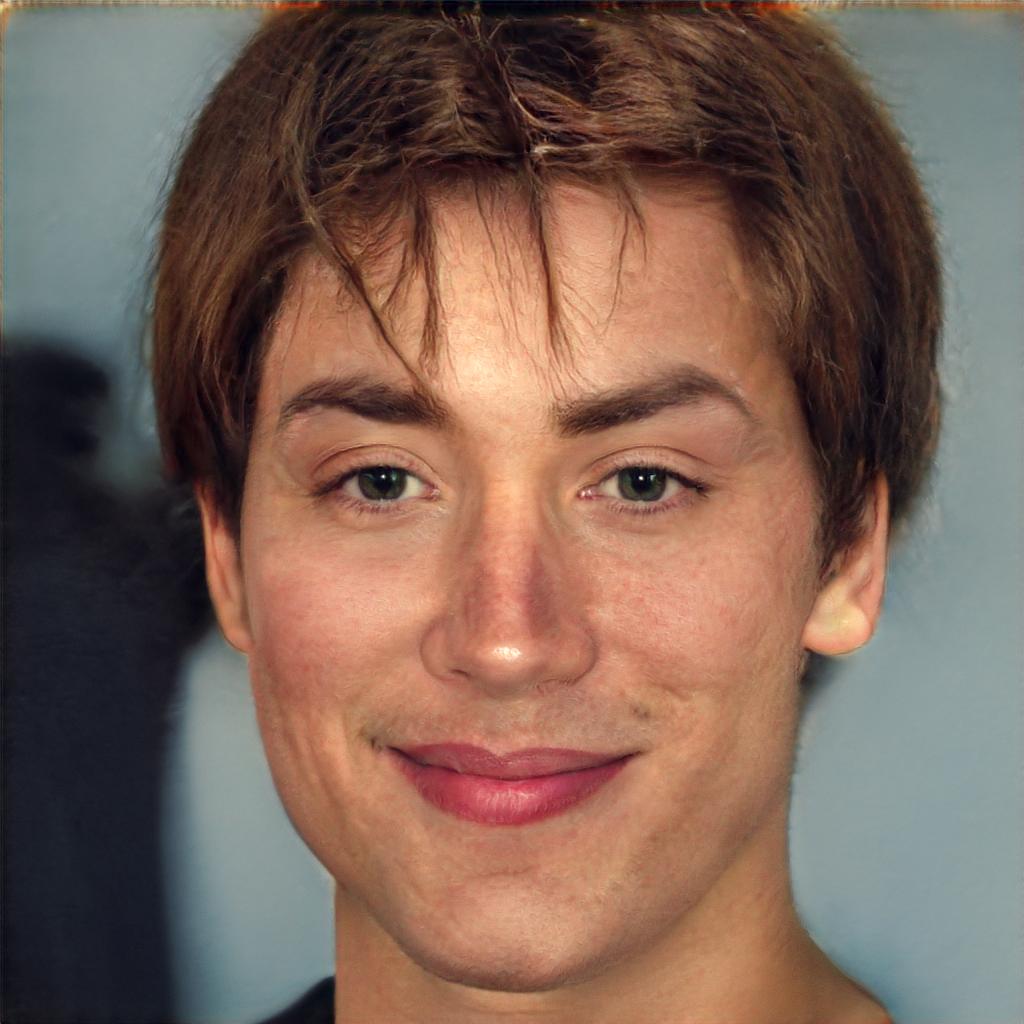}\vspace*{-3.5mm}\\
\begin{tikzpicture}\draw (0,0) -- (\linewidth,0);\end{tikzpicture}\vspace*{-0.5mm}\\%
\makebox[\hh]{\rotatebox[origin=l]{90}{\makebox[\h][c]{\hspace{-1mm}\normalsize{Fine from Source}}}}\hspace{0.5mm}%
\renewcommand{\seed}{seed845}
\includegraphics[height=\h]{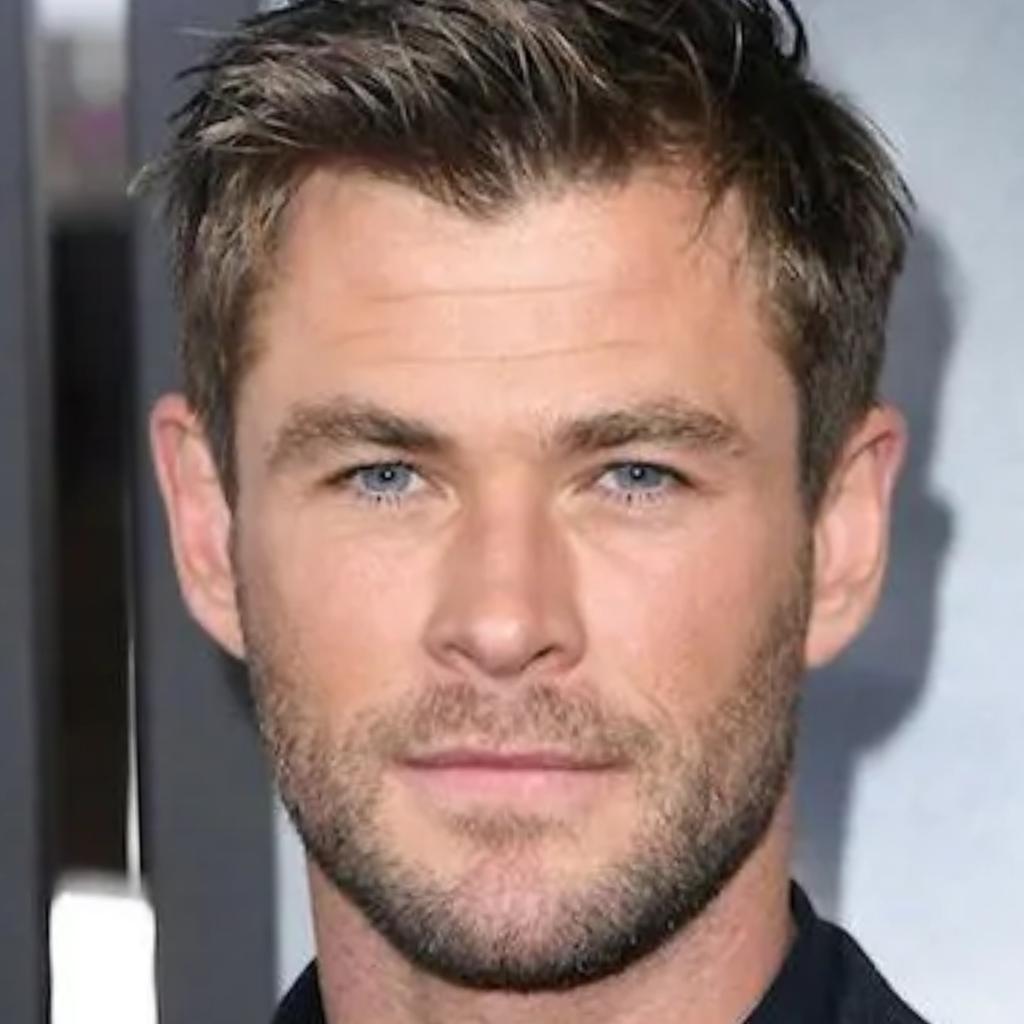}\hfill%
\raisebox{0mm}[0mm][0mm]{\makebox[0mm]{\hspace*{-1.5mm}\begin{tikzpicture}\draw (0,0mm) -- (0,-196.2mm);\end{tikzpicture}}}%
\includegraphics[height=\h]{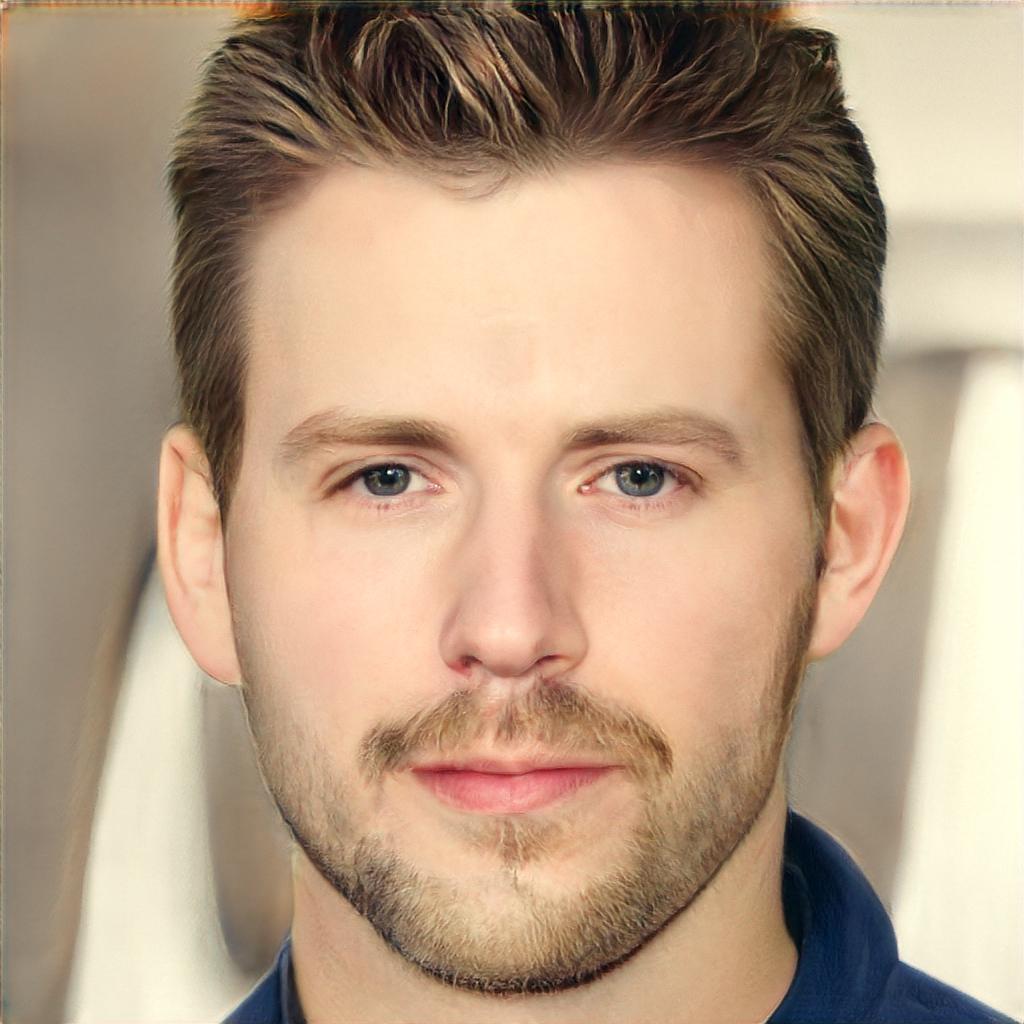}%
\includegraphics[height=\h]{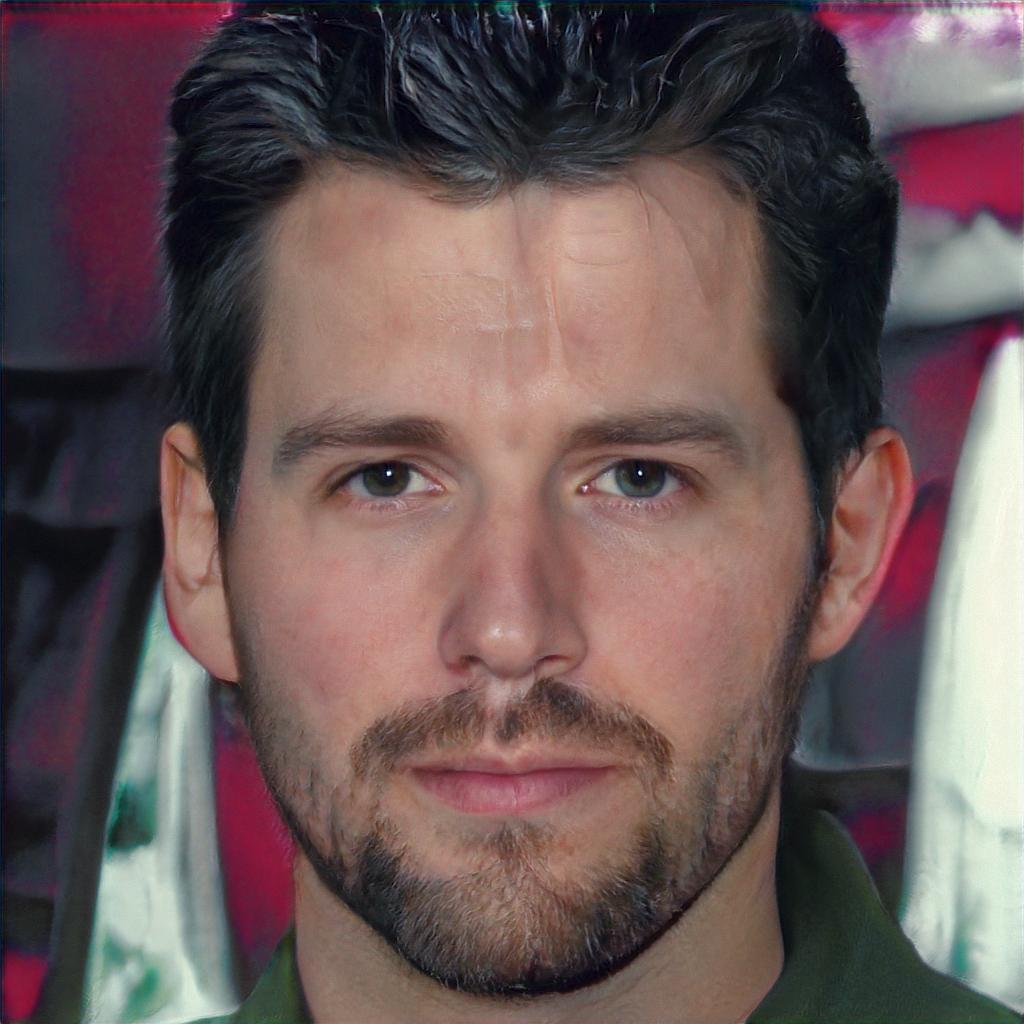}%
\includegraphics[height=\h]{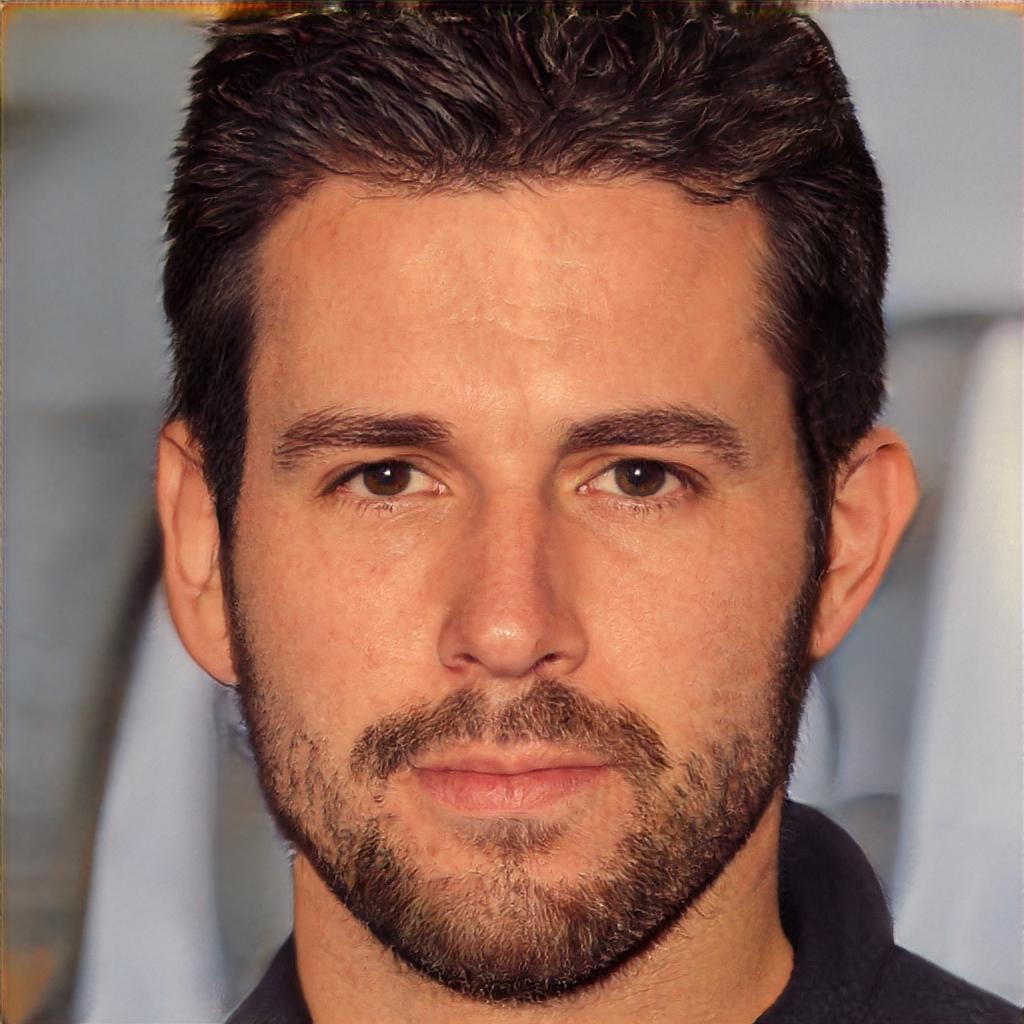}%
\includegraphics[height=\h]{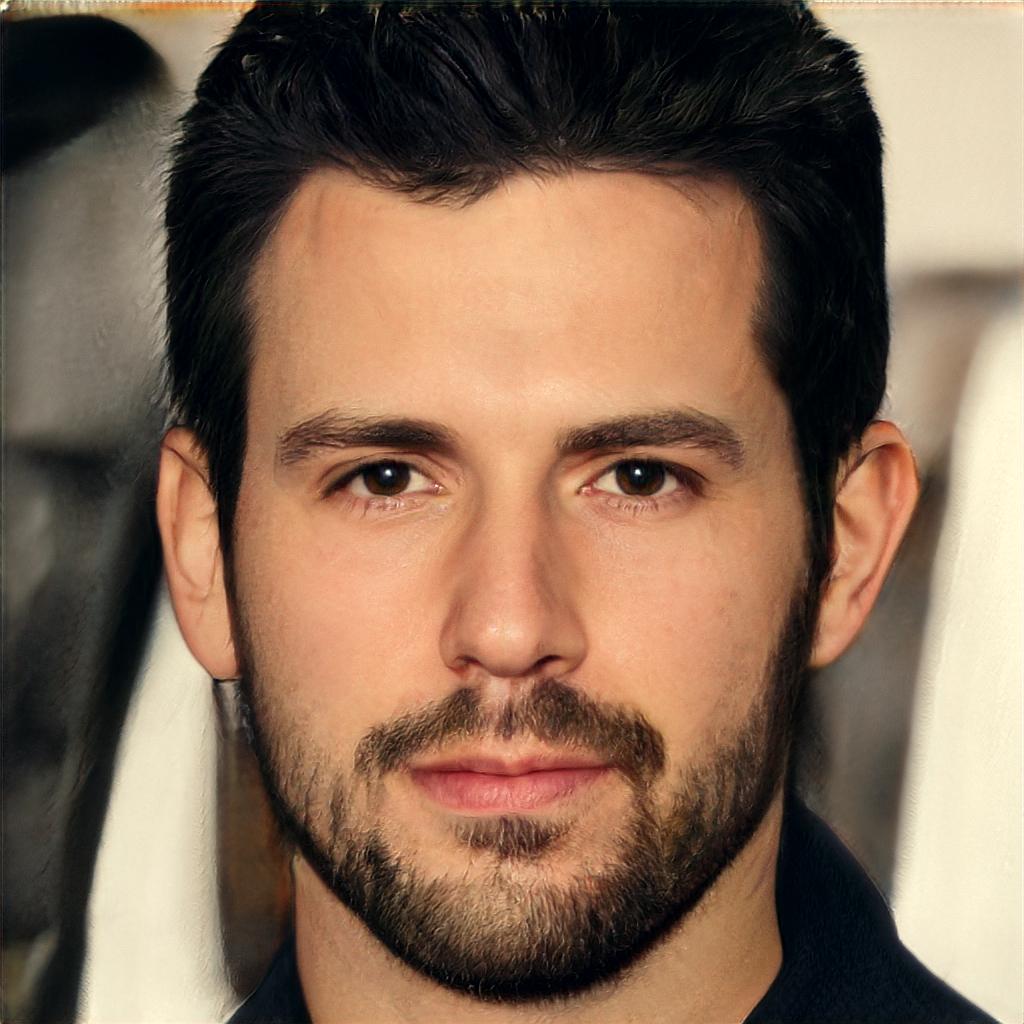}%
\includegraphics[height=\h]{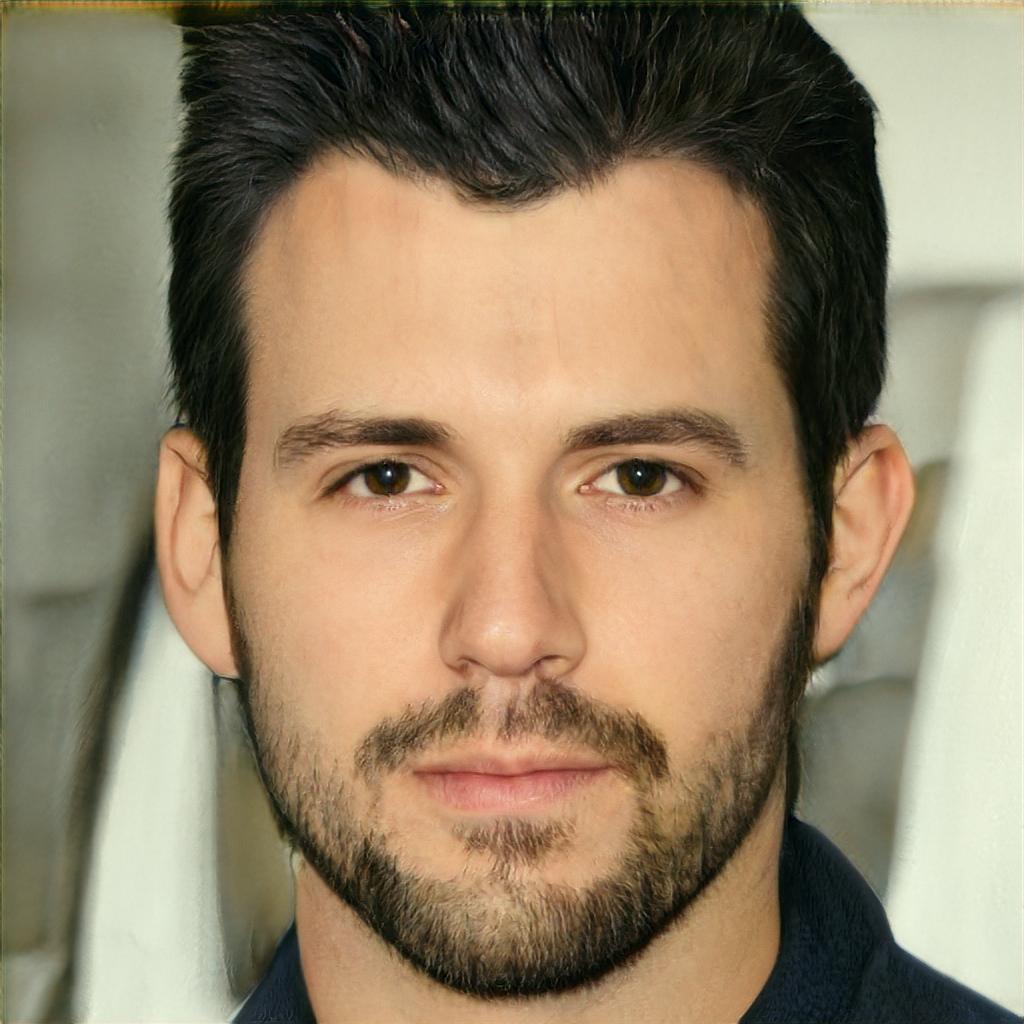}\\
\end{minipage}
\caption{\label{fig:stylemix}%
Two sets of real images were picked to form the Source set and the Destination set. The rest of the images were generated by copying specified subset of styles from the Source set into the Destination set. This experiment repeats the one from~\cite{Karras2019}, but with real images. Copying the coarse styles brings high-level aspects such as pose, general hair style, and face shape from Source set, while all colors (eyes, hair, lighting) and finer facial features resemble the Destination set. Instead, if we copy middle styles from the Source set, we inherit smaller scale facial features like hair style, eyes open/closed from Source, while the pose, and general face shape from Destination are preserved. Finally, copying the fine styles from the Source set brings mainly the color scheme and microstructure.
\vspace*{-1mm}
}
\end{figure*}

